\PassOptionsToPackage{table}{xcolor}
\documentclass[fleqn,10pt, table]{wlscirep}
\usepackage[table]{xcolor}
\usepackage[utf8]{inputenc}
\usepackage[T1]{fontenc}
\usepackage{booktabs}
\usepackage{float}
\usepackage{caption}
\usepackage{subcaption}
\usepackage{graphicx}
\usepackage{mwe}
\usepackage{nameref}
\usepackage{hyperref}
\usepackage{colortbl}
\usepackage{soul}
\usepackage[table]{xcolor}
\usepackage{multirow}
\usepackage[normalem]{ulem}
\usepackage{geometry}
\definecolor{lightyellow}{rgb}{1,1,0.6}

\sethlcolor{lightyellow}
\renewcommand\hl[1]{#1} % <<<<<<<<<<<<<<<<<<<<<<<<<<<<<<<<<<<<<

\definecolor{setup1}{HTML}{7B9FF9}
\definecolor{setup2}{HTML}{C0D4F5}
\definecolor{setup3}{HTML}{F2CBB7}
\definecolor{setup4}{HTML}{EE8468}

\def\code#1{\texttt{#1}}

\title{Exploring AI-based System Design for Pixel-level Protected Health Information Detection in Medical Images}

\author[1,*]{Tuan Truong}
\author[1]{Ivo M. Baltruschat}
\author[1]{Mark Klemens}
\author[1]{Grit Werner}
\author[1]{Matthias Lenga}
\affil[1]{Bayer AG, Berlin, Germany}

\affil[*]{tuan.truong@bayer.com}

\begin{abstract}
De-identification of medical images is a critical step to ensure privacy during data sharing in research and clinical settings. The initial step in this process involves detecting Protected Health Information (PHI), which can be found in image metadata or imprinted within image pixels. Despite the importance of such systems, there has been limited evaluation of existing AI-based solutions, creating barriers to the development of reliable and robust tools. In this study, we present an AI-based pipeline for PHI detection, comprising three key \hl{modules}: text detection, text extraction, and text analysis. We benchmark three models—YOLOv11, EasyOCR, and GPT-4o— across different setups corresponding to these \hl{modules}, evaluating their performance \hl{on two different datasets encompassing multiple imaging modalities and PHI categories. Our findings indicate that the optimal setup involves utilizing dedicated vision and language models for each module, which achieves a commendable balance in performance, latency, and cost associated with the usage of Large Language Models (LLMs). Additionally, we show that the application of LLMs not only involves identifying PHI content but also enhances OCR tasks and facilitates an end-to-end PHI detection pipeline, showcasing promising outcomes through our analysis.} 
\end{abstract}
\begin{document}
\flushbottom
\maketitle
\thispagestyle{empty}
\clearpage
\section*{Statements and Declarations}
\subsection*{Funding}
The authors declare that no funds, grants, or other support were received during the preparation of this manuscript.

\subsection*{Competing interest}
The authors declare they were employed by Bayer AG during the time of the research. No other financial or non-financial interests related to the work submitted for publication are disclosed.

\subsection*{Author contributions}
All authors contributed to the conception and design of the study. Tuan Truong was responsible for preparing materials, as well as for data collection and analysis. Ivo Baltrulschat and Matthias Lenga provided valuable input through discussions and contributed to the analysis of the experimental results. Mark Klemens and Grit Werner offered insights through their clinical discussions. All authors have read and approved the final manuscript.

\subsection*{Ethics approval}
This research study was conducted retrospectively using human subject data made available in open access. Ethical approval was not required, as confirmed by the license attached to the open-access data.

\subsection*{Consent to participate}
Not applicable 

\subsection*{Consent to publish}
Not applicable

\subsection*{Data availability}
The datasets used for training and evaluation of the pipeline are available upon request.

\subsection*{Acknowledgements}
The authors would like to thank the Bayer team of the AI Innovation Platform for providing computing infrastructure and technical support. Furthermore, we would like to thank Subrata Bose for fruitful discussions and valuable feedback.

\clearpage
\section{Introduction}

Protected Health Information (PHI) is health-related data that can identify an individual, often found in medical records, images, or any documents and data elements generated during participation in healthcare services. Before sharing data, e.g., for research purposes, it is crucial to de-identify or pseudonymize it to minimize the risk of re-identification. This de-identification step is mandatory for studies to comply with the Health Insurance Portability and Accountability Act (HIPAA) \cite{portability2012guidance} and the General Data Protection Regulation (GDPR) \cite{EuropeanParliament2016a}, and falls under the responsibility of the data processor. \hl{Furthermore, fully de-identified datasets enable broader and unrestricted use, facilitating collaborative research and benchmarking without the need for complex and time-consuming processes to obtain patient authorization and approval.} 

De-identification is a complex process that involves many data types and formats, typically including a detection step followed by the redaction
of the identified PHI. Medical images are a rich source of PHI, which can be embedded as metadata (e.g., in DICOM headers) or burned-in text within pixel data. \hl{However, manually detecting PHI in medical images is infeasible for large datasets, especially those that are curated for machine learning applications. Thus, the development of an automated tool for recognizing PHI is not just advantageous but essential for maintaining compliance and ensuring the privacy and security of sensitive health information. By leveraging automation, researchers and healthcare professionals can streamline the de-identification process, safeguarding patient privacy while enabling valuable insights from healthcare data.} 

\hl{Nevertheless,} the automatic detection of pixel-level PHI is challenging due to the diversity in imprint content, locations, font sizes, color schemes, annotations, and machine overlays. 
As the overall volume of medical data increases, so does the risk of incomplete PHI redaction, potentially leading to data breaches, which imply significant negative consequences for affected individuals and the healthcare sector, as well as high costs for responsible entities \cite{health2019cost}. The importance of reliable PHI detection at scale and automated image quality control in general is strongly emphasized by the latest medical image data platform development program launched by the Advanced Research Projects Agency for Health (ARPA-H) \cite{arpah_innovative_2024}. 

Previous works have employed Optical Character Recognition (OCR) technology to recognize text in medical images, leading to two primary approaches for handling burned-in text. The first approach \hl{leverages text detection or OCR models such as Tesseract {\cite{kay_tesseract_2007}} to localize burned-in texts and} removes all \hl{of them}, regardless of content {\cite{kline_medical_2023, langlois_open_2024, rempe_-identification_2024}}. While this method effectively eliminates any potential PHI, it also discards valuable information, such as measurements and markers. The second approach applies \hl{OCR models to extract burned-in texts and designs} heuristic rules to \hl{recognize PHI based on matching the extracted text to a defined PHI dictionary and subsequently redact them {\cite{macdonald_method_2024, monteiro_-identification_2017, vcelak_identification_2019}}}. \hl{While effective in structured settings, rule-based approaches are often tailored to specific imprint formats or language patterns and may struggle to generalize when faced with heterogeneous texts, ambiguous phrasing, or languages not covered by predefined rules.} In addition, fine-grained classification of PHI is often preferred to coarse classification, which categorizes information simply as names or dates. \hl{This is particularly needed in certain research contexts where information such as visit dates or study-specific identifiers may be retained under a data use agreement, as it is essential for study analysis.} As the volume and diversity of medical data continue to grow, advanced automatic solutions for PHI detection are essential, particularly for managing large datasets collected retrospectively. Therefore, a robust PHI detection tool must incorporate mechanisms that enable easy specification and adjustment of rules to meet the needs of different studies \hl{or country-specific guidelines}.

\begin{figure}[hpt]
    \centering
    \includegraphics[width=\textwidth, trim={1cm 5cm 0cm 4cm}, clip]{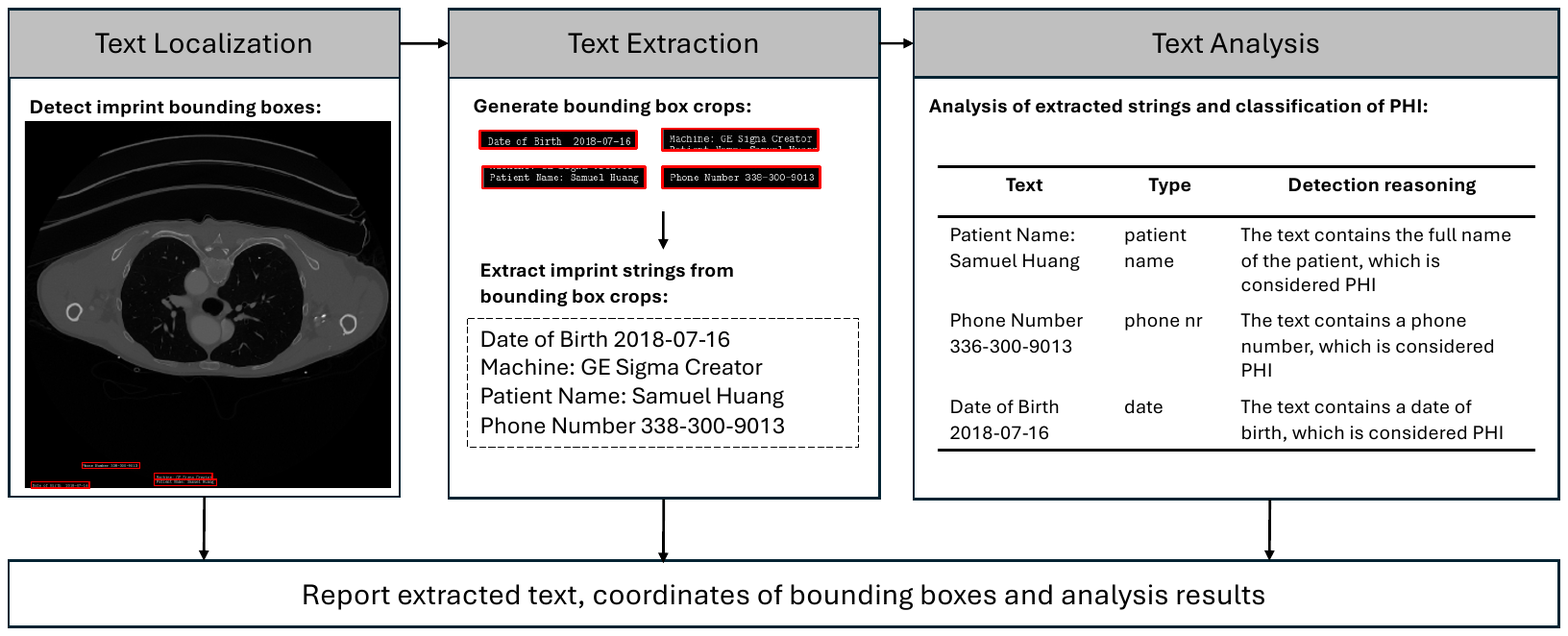}
    \caption{Pixel-level PHI detection pipeline. The pipeline consists of three components: text localization, text extraction, and text analysis. First, the bounding box coordinates of imprints are detected. Then, the text extraction module transforms the pixel-level imprints into machine-encoded texts. In the last step, the text analysis module identifies the PHI content.}
    \label{fig:pipeline}
\end{figure}

\begin{figure}[hpt]
    \centering
    \includegraphics[width=\textwidth, trim={0cm 6cm 0cm 4cm}, clip]{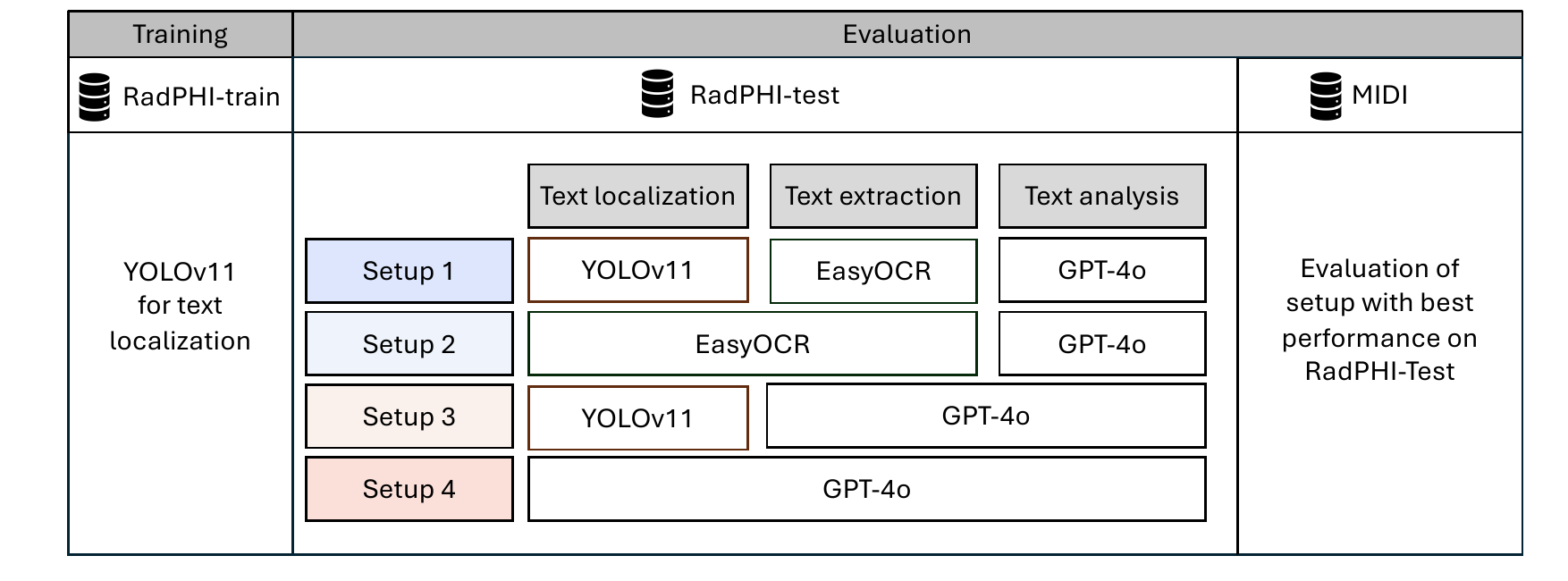}
    \caption{Proposed experimental framework for benchmarking different PHI detection pipelines. We employ three synthetic pixel-level PHI datasets: RadPHI-train for fine-tuning the text localization module, RadPHI-test for evaluating different combinations of YOLOv11, EasyOCR, and GPT-4o, and MIDI for further validation of the optimal setup derived from the RadPHI-test assessments}
    \label{fig:exp_setup}
\end{figure}

Our study focuses on AI-based approaches for automatically detecting PHI as burned-in pixels. \hl{Current commercial tools that address this need include MD.ai {\cite{mdai}}, Google {\cite{gcloud_healthcare}}, John Snow Labs {\cite{johnsnowlabs}}, and Glendor {\cite{noauthor_glendor_nodate}}. Additionally, Microsoft provides an open-source tool called Presidio {\cite{microsoft_presidio}} that serves similar purposes.} These tools commonly integrate two key components: a vision component that detects and converts burned-in texts into machine-encoded strings, and a language component that identifies any PHI content in the extracted texts \cite{jahan_comprehensive_2024}. To achieve this, OCR models are utilized for text recognition, while name-entity recognition (NER) models are employed to classify the PHI content. Recently, large language models (LLMs) have become the preferred candidates for the language component due to their flexible configurability and ability to understand complex language context in NER tasks, which has proven helpful in detecting PHI in biomedical text \cite{jahan_comprehensive_2024}. Despite some general remarks concerning the deployment of these tools \cite{clunie_summary_2024, clunie_summary_2024-1}, particularly on the non-deterministic behavior and cost of LLMs, there is a lack of comprehensive studies dissecting the setup of vision and language models within a PHI detection pipeline. In addition, as multimodal LLMs with vision capabilities are being released today \cite{openai_gpt4o_nodate}, the question remains whether they can be used to recognize PHI through end-to-end processing of images. 
To address the above concerns, we make the following contributions in our study:
\begin{itemize}
    \item We propose an end-to-end PHI detection system comprising three modules: text localization to identify text in images, text extraction to convert pixels into machine-readable text, and text analysis to detect PHI content using advanced language models (Figure~\ref{fig:pipeline}).
    \item \hl{We combine three models (i.e., YOLOv11, EasyOCR, and GPT-4o) corresponding to three previously mentioned modules in four different setups of a PHI detection pipeline (Figure~{\ref{fig:exp_setup}}) and compare their performance in an extensive quantitative study. In each setup, a model can assume one or multiple roles. For example, EasyOCR can perform not only text extraction but also text localization, and GPT-4o can perform end-to-end PHI detection instead of only analyzing the semantic content of extracted texts.}
    \item \hl{By leveraging publicly available medical imaging data and the HIPAA guidelines, we create two benchmark datasets comprising 1000 and 200 images, respectively. These datasets encompass various radiology modalities, including X-Ray, CT, PET, MRI, bone scintigraphy, mammography, and ultrasound. For both datasets we provide detailed annotations to support development and benchmarking within the research community.}
    \item Based on an extensive quantitative analysis, we provide insights into the optimal pipeline setup, considering performance, latency, and inference cost in terms of the number of generated tokens. Specifically, we explore the full capabilities of GPT-4o for both vision and language tasks, and analyze the trade-offs involved in utilizing it partially or entirely in the PHI detection pipeline.

    % \item We experiment with different module setups where the models can be assigned to their non-standard roles, and evaluate performance of the entire system on the simulated PHI dataset.
\end{itemize}

\section{Datasets}
\label{sec:dataset}
For the study at hand, we generate and utilize three distinct datasets:
RadPHI-train as our training dataset, and two evaluation datasets, RadPHI-test and MIDI. RadPHI-train and RadPHI-test are created by overlaying synthetically generated imprints on curated medical images across various modalities. MIDI is created by overlaying information stored in DICOM tags, which may contain PHI, utilizing an industry standard DICOM viewer. The RadPHI-train is used to develop our text detection model, while the RadPHI-test is employed to evaluate different configurations of the PHI detection pipeline. Additionally, the MIDI dataset is reserved as a hold-out test set to assess the optimal configuration identified through the RadPHI-test evaluation. In the upcoming sections, we will elaborate on the generation process for each of these datasets.

\subsection{PHI definition}
\label{phi-def}
For data simulation and evaluation, we define a list of PHI and non-PHI items based on the HIPAA guidelines \cite{portability2012guidance} as detailed in Table \ref{tab:phi-overview}. \hl{The six PHI categories include date, general identifier, patient name, address, phone number, and email}. We group all identification numbers into a group of identifiers, including patient ID, insurance number, social security number, and any number or series that can be used to identify an individual. \hl{The ten non-PHI categories include age < 90, gender, height, weight, examination type, hospital, marker, scanner, diagnosis, and imaging personnel.}

\begin{table}[htp]
    \centering
    \begin{tabular}{l p{7cm}l p{5cm}}
         \toprule
         \textbf{Category} & \textbf{Description} & \textbf{Classification} & \textbf{Example}\\
         \midrule
         Date & All elements of dates related to an individual, including birth date, admission date, discharge date, \hl{and all ages over 89 years, as well as all elements of dates indicative of such age} & PHI & \code{DOB 01-01-2023}\\
         \midrule
         Identifier & All identifiers related to an individual, such as patient identifier, insurance number, and medical record number & PHI & \code{Patient ID: 0000.0001}\\
         \midrule
         Patient name & Full name or initials of patient names & PHI & \code{Pat. Name: John Doe}\\ 
         \midrule
         Address & Partial or full address of the patient that contains any of the following: street address, city, state, postal code, and country & PHI & \code{123 Main St, Springfield, IL 62701, USA}\\
         \midrule 
         Phone number & Personal telephone number & PHI & \code{Contact 794-512-9544} \\ 
         \midrule
         Email & Personal email address & PHI & \code{Email: jane.smith@email.com}\\ 
         \midrule
         \hl{Age < 90} & \hl{Age that is below 90, as well as all elements of dates indicative of such age.} & \hl{Non-PHI} & \code{Age: 60}\\ 
         \midrule
         Gender & The gender of an individual, e.g., male, female, or diverse, or the abbreviations such as m/f/d & Non-PHI & \code{[M]}\\
         \midrule
         Height & Height measurement of the patient & Non-PHI & \code{Height: 165 cm}\\ 
         \midrule
         Weight & Weight measurement of the patient & Non-PHI & \code{Weight  103 kg}\\
         \midrule
         Examination type & The type of scan performed & Non-PHI & \code{Exam: CT Cholangiography}\\
         \midrule
         Hospital & General information about the hospital or imaging facility & Non-PHI & \code{Mayo Clinic Eau Claire}\\
         \midrule
         Marker & Text indicating anatomical markers or points of interest & Non-PHI & \code{R POST L}\\
         \midrule
         Scanner & General information about the scanner type and scanner settings & Non-PHI & \code{Philips Ingenia 3.0T}\\ 
         \midrule
         Diagnosis & General diagnosis, comments or indications made by doctors & Non-PHI & \code{Diagnosis: Fibrosis}\\ 
         \midrule
         Imaging personnel & Information about the medical staff involved in the scan, such as names of the radiologist, technician, operator, or referring physician who orders, acquires, performs, or interprets the scan & Non-PHI & \code{Indicated by John Moore}\\
         \bottomrule
    \end{tabular}
    \caption{Imprint categories and examples. There are 16 categories, six of which are classified as PHI.}
    \label{tab:phi-overview}
\end{table}

\subsection{Preprocessing of public datasets}
\label{ds-preprocessing}
For the generation of RadPHI-train and RadPHI-test datasets, we curate radiological images from well-known publicly available datasets. Each of them was preprocessed before we added simulated imprints. In the following, we briefly describe each dataset and the applied preprocessing.

\paragraph{Total Segmentator v2} \cite{wasserthal2023totalsegmentator} This dataset was originally published and used in a segmentation model for major anatomic structures in CT images. We use the second version, which includes 1,228 CT examinations across 21 scanners. Each volume is resampled to isotropic spacing of 0.45 mm across all three planes. Afterwards, four 2D axial slices are uniformly sampled per volume, which makes a total of 4,896 2D images. The images are min-max normalized to 8-bit format with the minimum value being random between the 0-th and 10-th percentile of the image intensities.

\paragraph{BS-80K} \cite{huang2022bs} This dataset includes 82,544 radionuclide bone scan images of 13 body regions designed for classification and object detection tasks. Experts de-identified and annotated the images for classification and object detection tasks. For our experiments, we select subsets of whole-body images in anterior and posterior views. Due to the small size of whole-body images, we create image collages of three to five randomly repeated images. This approach mimics a bone scan viewer, where multiple views of a subject are displayed side by side for comparison. To enhance the diversity of bone scan variants, backgrounds are randomly flipped. The processed images are then saved in 8-bit format.

\paragraph{ChestX-ray8} \cite{wang2017chestx} This dataset contains 108,948 frontal X-ray images with nine disease labels parsed from radiological reports. Many images contain physical markers indicating left and right direction. To avoid confusion with added text imprints, we use an object detection model to identify these markers and apply center cropping to remove them. After cropping, the images are saved in 8-bit format. 

\paragraph{BRATS} \cite{antonelli2022medical} This dataset is part of the Medical Segmentation Decathlon and includes 750 brain MRI images from four sequences: native T1, post-contrast T1, T2-weighted, and T2-FLAIR. All images are resampled to isotropic spacing of 1 mm. For each subject, a random sequence is selected, and four 2D axial slices are uniformly sampled. The images are then min-max normalized to 8-bit format, following a similar approach to the Total Segmentator v2 dataset. 

\subsection{RadPHI-train}
\label{training-ds}
The main use of the RadPHI-train dataset is to enhance the robustness of the text localization module (see Section \ref{sec-methods}) against diverse radiological image backgrounds, varying numbers of imprints, and different imprint representations.
This dataset is generated based on images from TotalSegmentator, BS-80K, and ChestX-ray8, applying the preprocessing techniques outlined in Section \ref{ds-preprocessing}. Additionally, we simulate imprints on synthetic backgrounds, incorporating plain grayscale colors along with or without multiple boxes in various shades. The text localization module is trained to manage diverse representations of the imprint rather than focusing on its text content. To achieve this, we continuously simulate the imprints using different fonts, sizes, colors, and locations. These imprints may overlap with each other or with anatomical structures. Although such overlaps are not typically seen in real-world data, we intentionally include them as challenging cases for the model to adapt to. A sample from this dataset is shown in Figure~\ref{fig:yolo-training-data} (left) in the Supplementary Material. In total, we generate 6,000 images, which are then divided into training and validation sets using an 80\%-20\% split. 

\subsection{RadPHI-test}

\begin{figure}[hpt]
    \centering
    \includegraphics[width=\linewidth, trim={0cm 3.5cm 0cm 3cm}, clip]{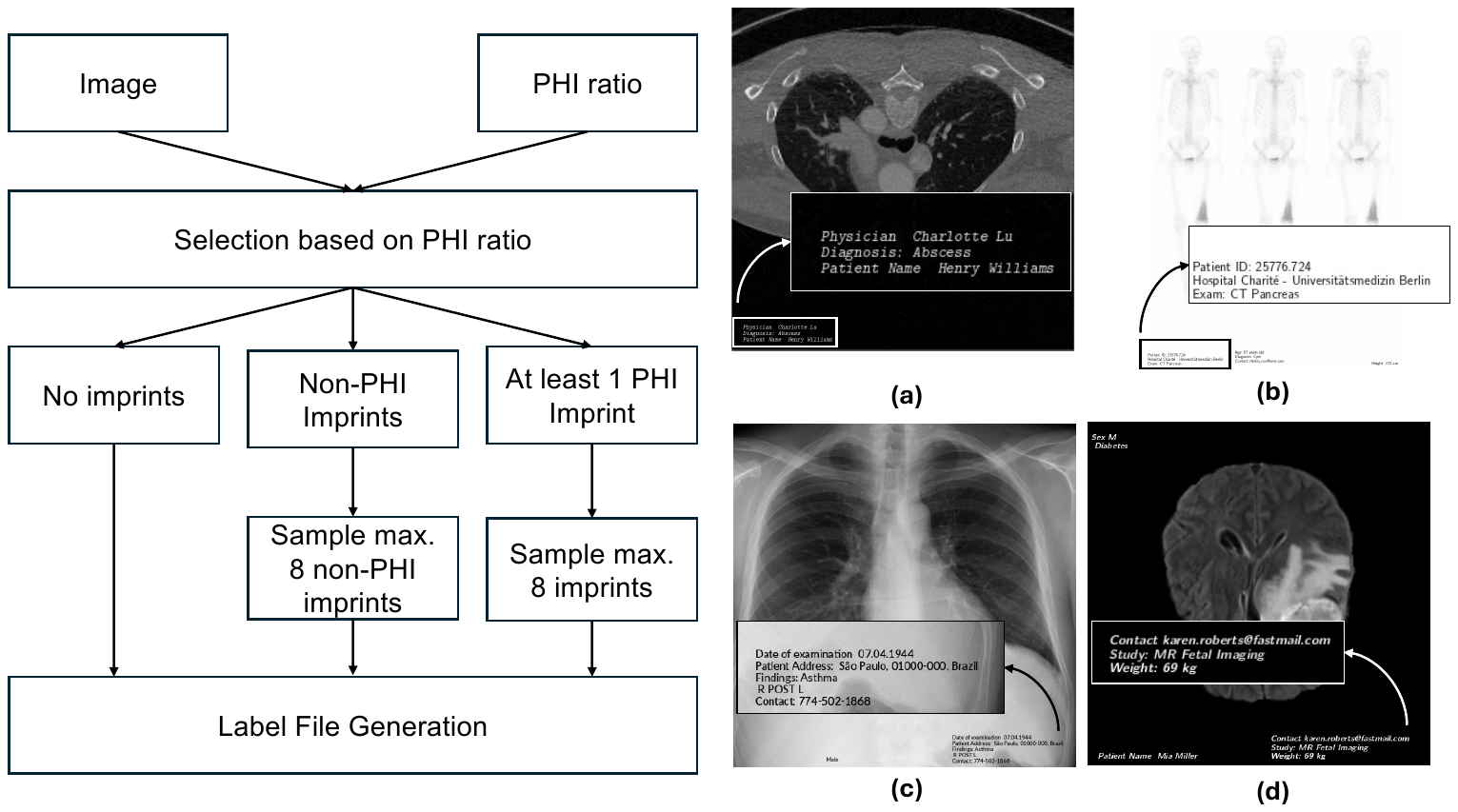}
    \caption{Imprint simulation pipeline and examples. (Left) Given the input PHI ratio for each dataset, each image can have no imprints, non-PHI imprints, or at least one imprint that contains PHI. The maximum of imprints per image is eight. (Right) Examples of images with imprints across four modalities: (a) Whole body CT (b) Whole body bone scan (c) Chest X-ray (d) Brain MRI.}
    \label{fig:imprints-generation}
\end{figure}
\begin{figure}
    \centering
    \includegraphics[width=\textwidth, trim={1cm 3cm 3cm 5cm}, clip]{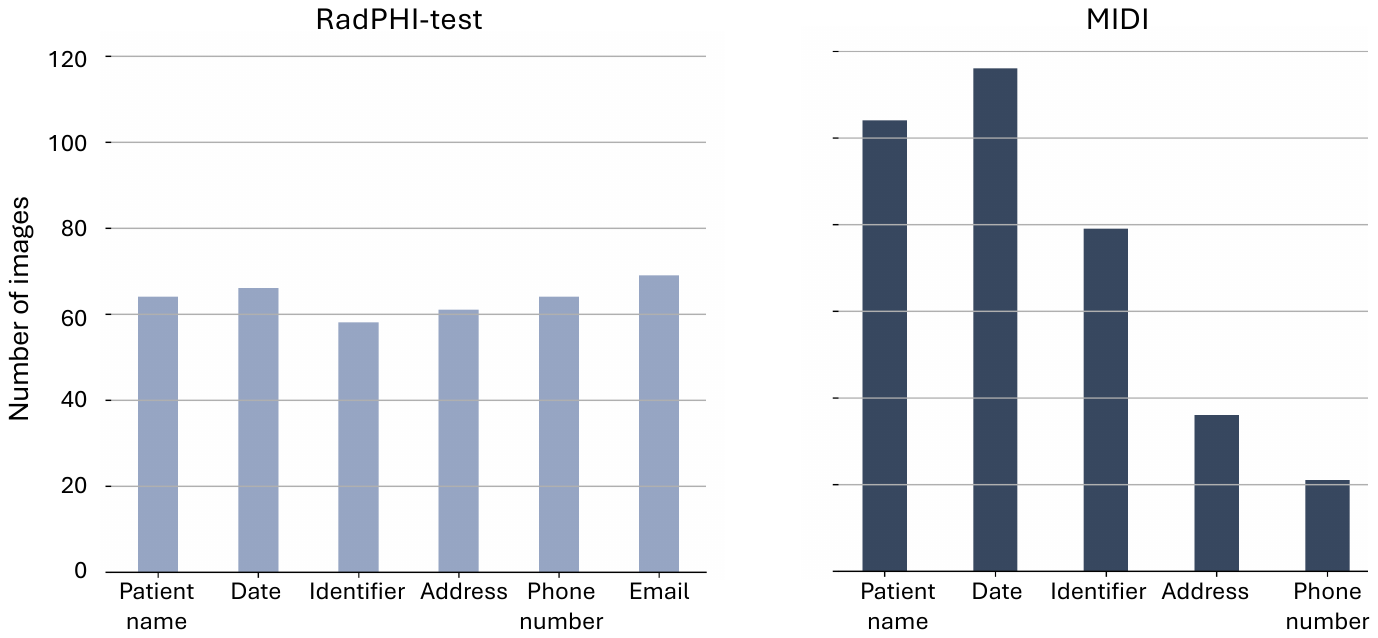}
    \caption{\hl{Distribution of PHI categories in the RadPHI-test and MIDI datasets. In the RadPHI-test dataset, each image can have a maximum of one imprint per category. In contrast, the MIDI dataset allows for multiple imprints of the same category in each image. We compute the distribution of PHI categories in the MIDI dataset at the image level, meaning that regardless of the number of imprints for a particular category, it is counted as one.}}
    \label{fig:cat-stats}

\end{figure}
\begin{figure}
    \centering
    \includegraphics[width=\textwidth, trim={1cm 2.5cm 5cm 2cm}, clip]{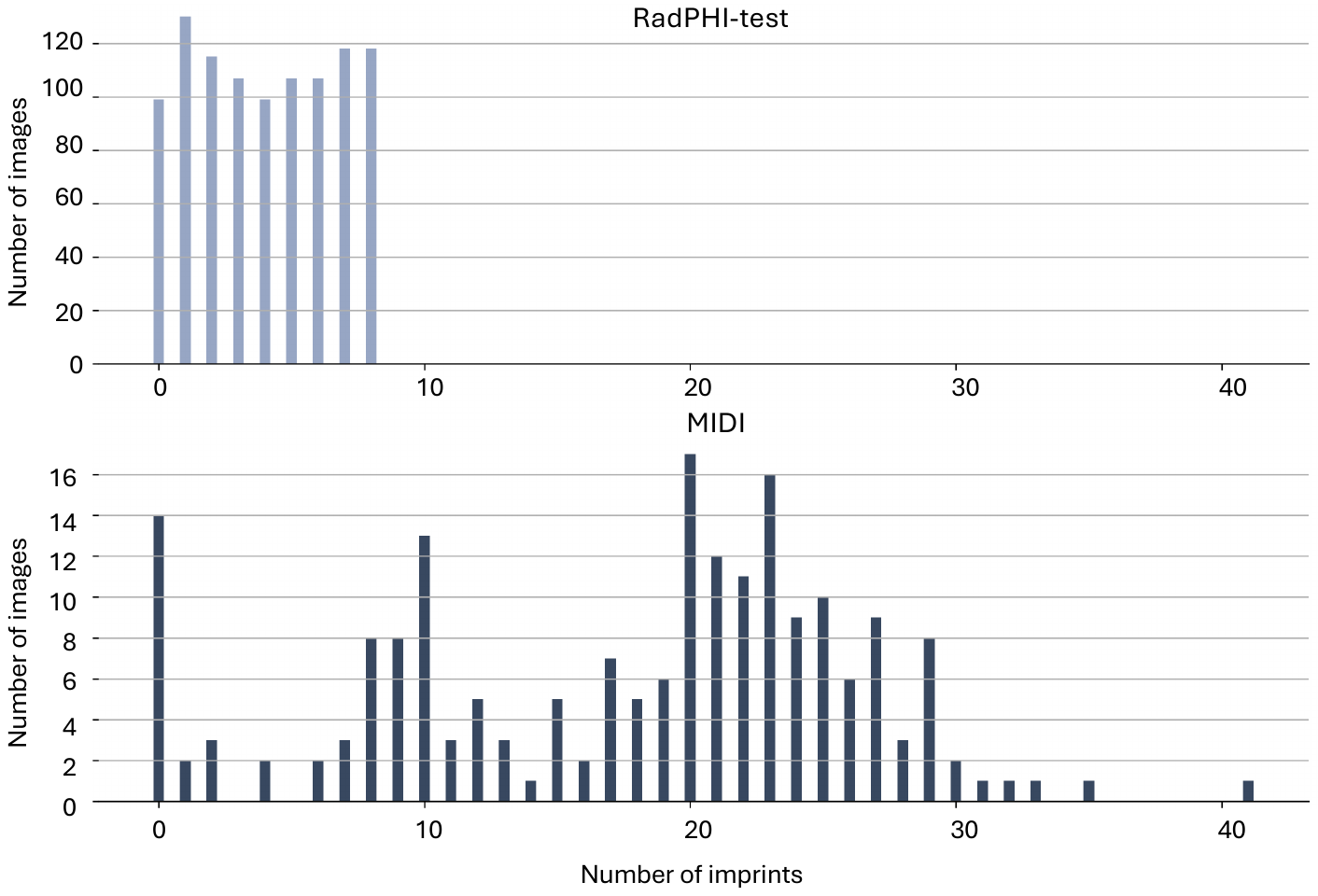}
    \caption{Distribution of PHI and non-PHI imprints in RadPHI-test and MIDI datasets. In RadPHI-test, the number of imprints is limited to eight. In contrast, MIDI has significantly more imprints, correlating to the various DICOM tags and modality-specific imprints randomly displayed on the viewer.}
    \label{fig:imprint-stats}

\end{figure}

The RadPHI-test is designed to evaluate various configurations of our PHI detection pipelines. To achieve this, we create a dataset that simulates real-world image imprints as closely as possible. The imprint simulation process is illustrated in Figure~\ref{fig:imprints-generation} (left). Based on the input ratio of PHI imprints for the simulated dataset, we determine for each image whether (1) no imprints are added, (2) only non-PHI imprints are added, or (3) at least one imprint includes PHI. Each image can contain a maximum of eight imprints. For each imprint category, as listed in Table~\ref{tab:phi-overview}, we generate imprints in the format \code{<accompanying text> <separator> <main text>}. The accompanying texts consist of signal words that indicate the category of the main text. The separator can be a comma \code{,} or a space. For example, in the text \code{Patient Name: John Doe}, \code{ Patient Name} serves as the accompanying text, \code{:} is the separator, and \code{John Doe} is the main text. In some cases, we randomly omit the accompanying text, such as with identifiers where imprints may consist solely of a sequence of numbers. This omission tests the language model's ability to infer the underlying imprint category robustly. After adding the imprints, a corresponding label file is generated. This label file contains the coordinates of each imprint in the image, along with its associated PHI class and sub-class. These labels are used for subsequent evaluation. The final dataset used to evaluate the PHI pipeline consists of 1,000 images distributed equally across four modalities. Of these, 850 images ($85 \%$) contain at least one PHI imprint. The distribution of the 16 imprint categories and the number of imprints per image are depicted in Figure~\ref{fig:cat-stats} and Figure~\ref{fig:imprint-stats}.

\subsection{MIDI}
\begin{figure}
    \centering
    \includegraphics[width=0.7\textwidth, trim={0cm 2.5cm 5cm 2cm}, clip]{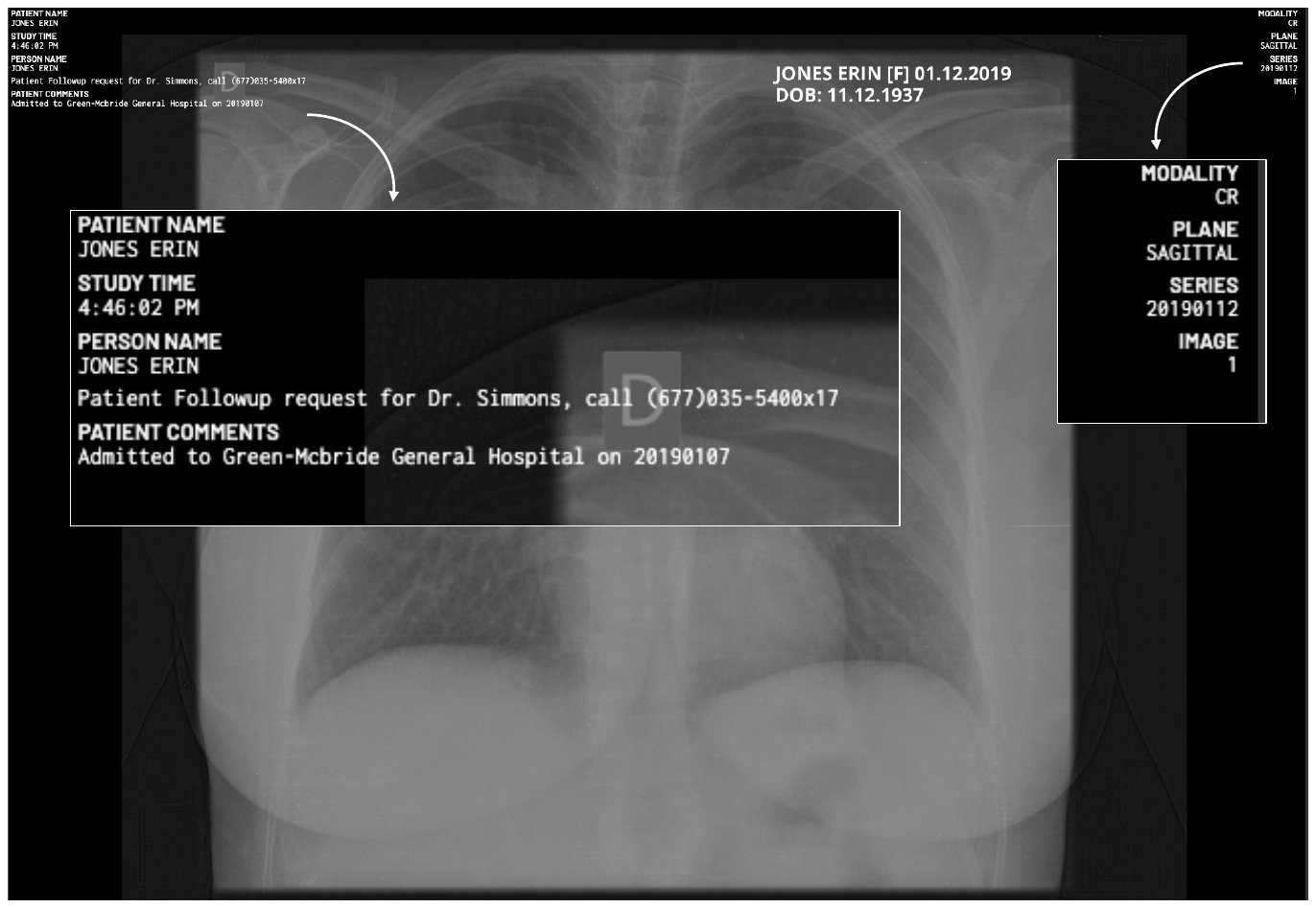}
    \caption{\hl{An example of an image from the MIDI dataset featuring PHI overlays. The same PHI element, such as dates, may appear in various contexts, including patient comments and image series. Besides overlays added by the DICOM viewer (left and right), there are burn-ins by the organizer of the MIDI-B challenge (middle).}} 
    \label{fig:midi-data}
\end{figure}

The MIDI dataset is designed to be a challenging hold-out dataset with realistic imprint visualization and placement via a modern DICOM viewer rendering.
The image data is curated from the validation set of the 2024 Medical Image De-Identification Benchmark (MIDI-B) challenge \cite{infosagebaseorg_midi-b_nodate}. This dataset originally consists of 228 DICOM images across multiple modalities, each containing synthetic PHI content embedded at both the DICOM header and pixel level. We utilize a DICOM viewer, specifically MD.ai \cite{mdai_viewer}, to overlay the DICOM tags onto the images. We randomly sample DICOM tags to ensure that the generated imprints represent all possible PHI categories, similar to the RadPHI-test dataset. After applying the overlays, we export the images from the viewer. The resulting images may include not only the DICOM tag overlays but also burn-ins by the challenge organizers. The final version of the dataset comprises 200 images categorized into five PHI types. The category Email is omitted because emails are neither burned in nor inserted into the header of the original dataset. In contrast to the RadPHI-test dataset, each image can have multiple imprints of the same PHI category. For instance, as illustrated in Figure~\ref{fig:midi-data}, dates may appear in various contexts, such as the study date, image series, or even the patient comments. \hl{We performed instance-level annotation of the images by generating coordinates for PHI instances along with their corresponding categories. This annotation process was carried out and validated by two independent annotators to ensure accuracy and reliability.}
Figure~\ref{fig:cat-stats} displays the distribution of PHI categories at the image level. Dates and patient names are among the most frequently occurring PHI elements in the dataset. While we intentionally limit the number of imprints to eight in RadPHI-test, the number of imprints in MIDI can be up to 41, with the majority between 10 and 30 per image (Figure~\ref{fig:imprint-stats}). 

\section{Methods}
\label{sec-methods}
The general design of the PHI detection pipeline comprises three major modules: text localization, text extraction, and text analysis (see Figure~\ref{fig:pipeline}). The text localization module can be an object detection model that identifies pixels containing text in the input image and returns bounding box coordinates. Once the text coordinates are determined, the text extraction module converts the text within these coordinates into machine-readable text, using OCR algorithms. The machine-encoded text is then analyzed semantically by a language model, which classifies whether it contains any PHI. In the last step, the pipeline flags the image as containing PHI, prompting further review and potential redaction. \hl{In our study, we explore three distinct models: YOLOv11, EasyOCR, and GPT-4o. These models are configured in various combinations to fulfill the roles of the three modules mentioned above. We will describe each model in detail and outline how they are integrated to perform the functions of text localization, text extraction, and text analysis.}

\paragraph{\hl{YOLOv11}}  YOLOv11 \cite{ultralytics_yolo11_nodate} is the latest version in the series of YOLO models \cite{redmon_you_2016}, offering superior performance in real-time object detection tasks. \hl{In our study, we employ YOLOv11 exclusively for the text localization module.} The YOLOv11 architecture is a single-shot object detector consisting of three primary components: a backbone, a neck, and prediction heads. The backbone extracts image features, which are then processed and aggregated across multiple scales in the neck. The final predictions are produced by classification and detection heads. The main improvement in YOLOv11 over the previous version is an optimized backbone and neck, which enhance the extracted features and enable more precise object detection. For our study, we use the small variant of YOLOv11 pre-trained on the COCO dataset and fine-tune it on the RadPHI-train dataset. The model is trained to predict bounding boxes around a single line of text in the image, as illustrated in Figure \ref{fig:pipeline}. In addition, we combine real medical images with synthetic images of diverse background colors and shapes to help the model remain robust across all modalities. We use the ultralytics package \cite{ultralytics_yolo11_nodate} to fine-tune the model with a fixed image size of 640x640 for 100 epochs. The other training hyperparameters are kept as default in the package. Apart from regular geometric augmentation techniques, the images undergo mosaic augmentation, where a new image is created based on patches of original images, as shown in Figure~\ref{fig:yolo-training-data} (bottom right). The mosaic augmentation helps the model to generalize better to diverse backgrounds and object locations.

\paragraph{\hl{EasyOCR}} EasyOCR \cite{noauthor_jaidedaieasyocr_nodate} is an open-source OCR package supporting over 80 languages. It comprises two primary modules: recognition and detection. The detection module, analogous to our localization module, identifies text regions in an image using the Character Region Awareness For Text detection (CRAFT) model \cite{baek_character_2019}. The recognition module \cite{baek2019STRcomparisons} extracts text by leveraging the Scene Text Recognition module, a multi-stage model that transforms input images, extracts features, aggregates contextual information, and finally outputs the characters.
\hl{We investigate EasyOCR under two distinct setups: (1) as a dedicated text extractor, and (2) as both a text localizer and a text extractor. For both setups,} we use EasyOCR as an out-of-the-box tool without re-training or fine-tuning its components. We keep the values of all parameters as default except for the low-bound text threshold. We reduced this threshold to 0.2 to include more space around detected texts by the CRAFT model, which we found helpful in avoiding missing texts as single letters like markers (e.g., L or R) or abbreviations of gender (e.g., M or F or D). The input to EasyOCR can be either a full image or a cropped image of the text region detected by the localization model, depending on the experiment (see Section~\ref{sec:experiment}).

\paragraph{\hl{GPT-4o}} GPT-4o \cite{openai_gpt4o_nodate} is OpenAI's flagship language model in the Generative Pre-trained Transformer (GPT) family. These models are pre-trained on extensive unlabeled text datasets to generate human-like responses. GPT models have demonstrated their versatility in information extraction and NER tasks, particularly in biomedical text analysis \cite{jahan_comprehensive_2024}. \hl{In our experiments, GPT-4o is configured to perform three distinct tasks: (1) text analysis, (2) text extraction, and (3) end-to-end image analysis. In the text analysis configuration, GPT-4o examines} the context of extracted texts from images in order to determine whether they contain PHI and which type of PHI is present. The extracted texts from the text extraction module are provided to GPT-4o as a list. The system prompt consists of the definition of PHI and non-PHI texts and instructions to evaluate the aggregated context of the input text rather than analyzing each line in isolation. \hl{When GPT-4o acts as a text extractor, the prompts are structured to ensure that it returns the text visible in each cropped region predicted by YOLOv11. In the end-to-end image analysis scenario, GPT-4o is presented with a full image, prompted to perform text recognition, and tasked with analyzing each imprint identified in the image. It's important to know that GPT-4o is not a dedicated object detection model, it does not return coordinates for the imprint. For the analysis task,} we provide examples in the prompt to make the model aware of ambiguous cases, following the few-shot prompting technique \cite{brown2020language}. For instance, \code{Age:} or \code{65} refers to a placeholder for age or hints at a number which could be age, but neither is considered PHI. To ensure consistent outputs, we constrain GPT-4o to produce a dictionary containing mandatory fields through OpenAI's pydantic tool, as outlined in Table \ref{tab:gpt-response}. If the PHI is detected, the \code{type} field specifies one of the eight PHI classes or \code{other} if the PHI category is not included in the provided list. If no PHI is detected, the text analyzer assigns \code{non-phi} to the \code{type} field. Additional fields include \code{raw\_text}, \code{reason}, and \code{language}, which store the raw extracted text, the justification for classifying the text as PHI or non-PHI, and the detected language of the original text, respectively. The temperature parameter is set to 0 to yield more deterministic responses.

\begin{table}[htp]
    \centering
    \begin{tabular}{c c p{8cm}}
         \toprule
         \textbf{Tag} & \textbf{Data type} & \textbf{Description}  \\
         \midrule
         \code{type} & \code{str} & One of \code{[date, identifier, patient\_name, address, phone\_nr, email, other, non-phi]}\\
         \midrule
         \code{raw\_text} & \code{str} & Original text extracted from the image \\
         \midrule
         \code{reason} & \code{str} & Reason why the extracted text is considered PHI or not \\ 
         \midrule
         \code{language} & \code{str} & Detected language \\
         \bottomrule
    \end{tabular}
    \caption{Structured output of the text analyzer GPT-4o. }
    \label{tab:gpt-response}
\end{table}

\section{Experiment}
\label{sec:experiment}
In this study, we assess various setups of a PHI detection pipeline by interchanging the roles of the presented models. Specifically, we study the four setups illustrated in Figure~\ref{fig:exp_setup}. The first setup employs three distinct models, each dedicated to one of the modules described in Section~\ref{sec-methods}.
In the second setup, we assess the robustness of EasyOCR's built-in object detection capabilities by using it for both text localization and extraction tasks. EasyOCR handles the detection of text regions and subsequently extracts the corresponding text. 
The third setup evaluates the language-vision capability of GPT-4o by assigning the text extraction step to it. In this configuration, GPT-4o receives cropped text regions, localized by the YOLOv11 model, performs OCR, analyzes, and classifies PHI content. 
The final setup explores GPT-4o as an end-to-end solution. GPT-4o takes full responsibility for all three pipeline steps: localizing text in the image, extracting the text, and analyzing it for PHI content. This setup, in particular, highlights a new use case where a single language model integrates vision and language tasks, demonstrating the potential for end-to-end applications.
For each of the introduced setups, we run the pipeline five times on the RadPHI-test dataset to assess the non-deterministic behavior of GPT-4o in generating responses. \hl{The final output of the PHI pipeline is evaluated at two levels: case and instance. At the case level, the focus is on whether the model can identify the presence of at least one PHI instance or a specific PHI class within the image. At the instance level, the evaluation checks whether the model accurately localizes every PHI instance or every instance of a specific class in the dataset. Performance is measured using recall, precision, false negatives (FN), and false positives (FP), which together provide insight into the model's sensitivity in identifying PHI. A summary of the evaluation framework can be found in Table~{\ref{tab:evaluation-scheme}}. 
In addition to the four pipeline variants, Microsoft's open-source tool Presidio {\cite{microsoft_presidio}} is included in the benchmark. Presidio is designed for de-identifying images using NER models and regular expression patterns. We create two custom categories for patient names and identifiers using regular expressions, while also utilizing other built-in categories such as date, address, phone number, and email. Since Presdio is deterministic, we run it only once on our data and evaluate its performance. Our comparison with Presidio serves as a baseline for evaluating our LLM-based approaches.}
The setup that performs best on the RadPHI-test dataset is then further evaluated on the MIDI dataset using the same evaluation metrics.

\begin{table}[htp]
    \centering
    \resizebox{\textwidth}{!}{%

    \begin{tabular}{c c p{6cm} c p{3cm}}
        \hline
        \textbf{Level} & \textbf{Evaluation type} & \textbf{Interpretation} & \textbf{Metric} & \textbf{Dataset} \\
        \hline
        Case & PHI presence & Did the model correctly identify the presence of at least one PHI instance in the image? & Recall, Precision, FN, FP & RadPHI-test, MIDI \\
        \hline
        Case & PHI class presence & Did the model correctly identify the presence of at least one PHI instance of class X in the image? & Recall, Precision, FN, FP & RadPHI-test, MIDI \\
        \hline
        Instance & PHI presence & Did the model correctly localize every PHI instance in the dataset? & Recall, Precision, FN, FP & RadPHI-test\textsuperscript{(*)}, MIDI \\
        \hline
        Instance & PHI class presence & Did the model correctly localize every PHI instance of class X in the dataset? & Recall, Precision, FN, FP & RadPHI-test\textsuperscript{(*)}, MIDI \\
        \hline
    \end{tabular}
    }
    \caption{\hl{Evaluation framework for PHI detection. The outputs are evaluated at two levels: Case-level, which considers whether the model detects the presence of at least one PHI instance or a specific PHI class in an image; and Instance-level, which evaluates the model’s ability to accurately localize each individual PHI instance or every instance of a specific PHI class across the dataset. (*) Instance-level evaluation is not performed in Setup 2 and Setup 4 because the coordinates are either not compatible to match with the ground-truth annotation (setup 2) or cannot be predicted (setup 4).}}
    \label{tab:evaluation-scheme}
\end{table}

% \begin{table}[]
%     \centering
%     \begin{tabular}{c|c}
%         TP &  PHI class X in GT and in detection result\\
%         FP &  PHI class X not in GT but in detection result\\
%         TN &  PHI class X not in GT and not in detection result\\
%         FN &  PHI class X in GT but in not detection result
%     \end{tabular}
%     \caption{for computation of end-2-end metrics}
%     \label{tab:my_label}
% \end{table}

% - to measure in more detail LLM effets:

% \begin{table}[]
%     \centering
%     \begin{tabular}{c|c}
%         TP (type1) &  PHI class X in GT and in detection result\\
%         TP (type2) &  PHI class X in GT, not detected in OCR stage but LLM coincidentally adds class to detection result\\
%         FP &  PHI class X not in GT but in detection result (due to wrong OCR or hallucination)\\
%         TN (type1) &  PHI class X not in GT, LLM does not detect class X\\
%         TN (type2) &  PHI class X not in GT, LLM hallucinates that class is present but coincidentally classifies it as non-PHI\\
%         FN (type1) &  PHI class X in GT, detected by LLM but wrongly classified\\
%         FN  (type2) &  PHI class X in GT, not dectect by LLM at all
%     \end{tabular}
%     \caption{more detailed breakdown}
%     \label{tab:my_label2}
% \end{table}

% - is there probability / confidence score returned nativly by LLM that can we can use for AUROC analysis?

\section{Results}
\subsection{Rad-PHI dataset}
\begin{figure}
    % \captionsetup[subfigure]{labelformat=empty}

    \centering
    \begin{subfigure}[b]{0.45\linewidth}
        \centering
        \includegraphics[width=\textwidth, trim={0cm 2cm 6.5cm 2.7cm}, clip]{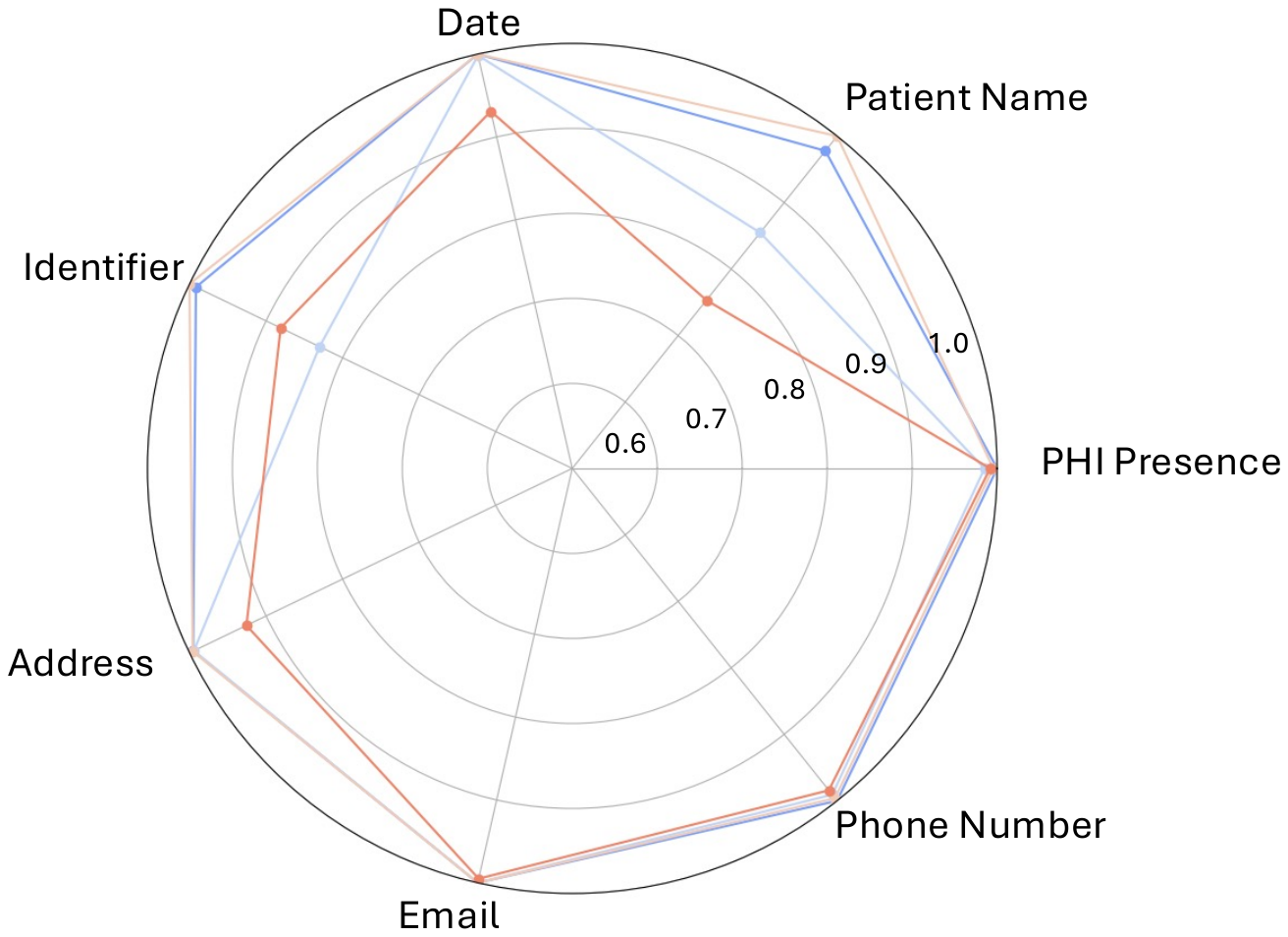}
        \caption{Case-level precision}
        \label{fig:case-precision}
    \end{subfigure}
    % \hfill
    \begin{subfigure}[b]{0.45\linewidth}
        \centering
        \includegraphics[width=\textwidth, trim={0cm 2cm 6.5cm 2.4cm}, clip]{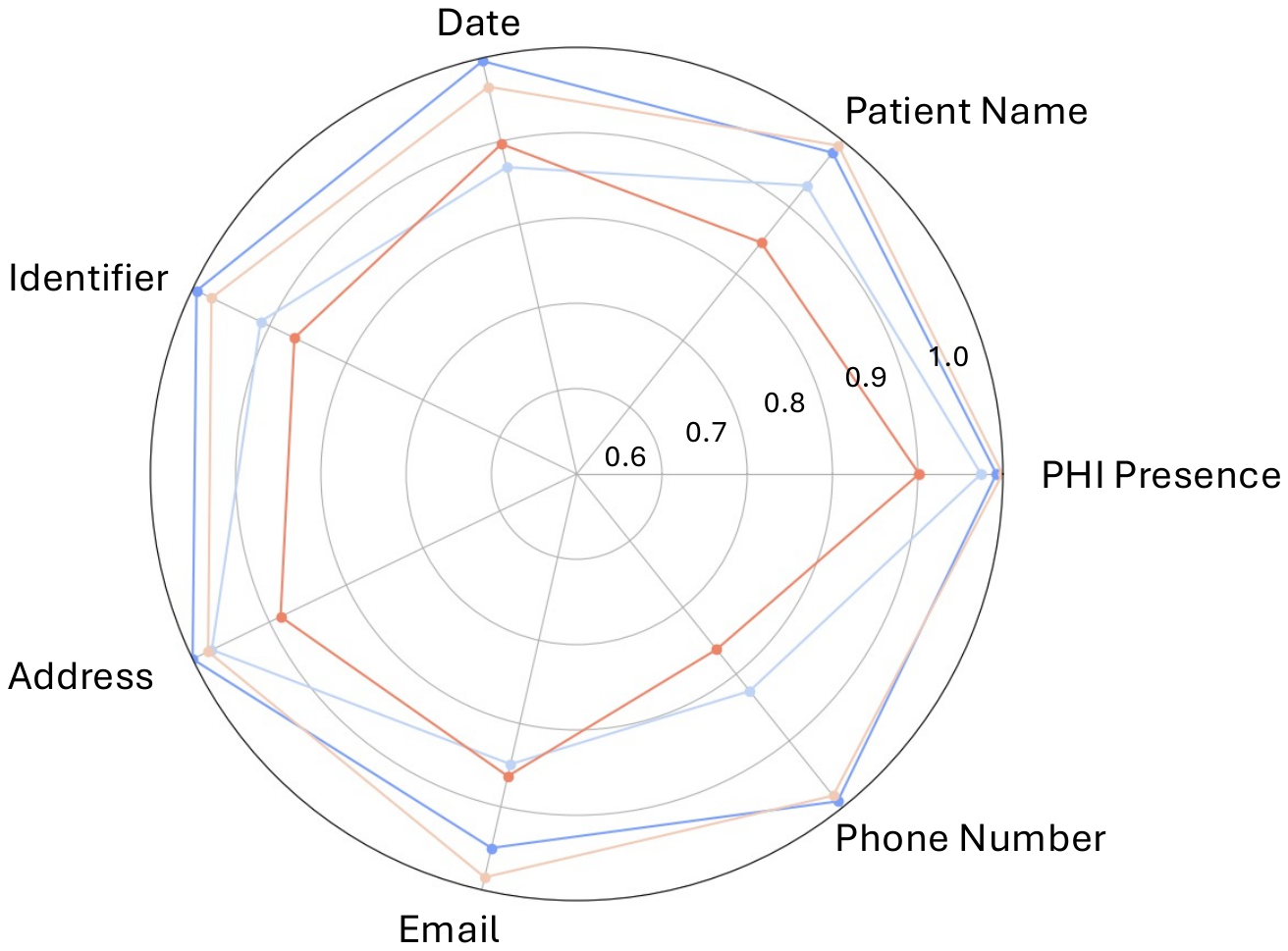}
        \caption{Case-level recall}
        \label{fig:case-recall}
    \end{subfigure}
    \vskip\baselineskip
    \begin{subfigure}[b]{0.45\linewidth}
        \centering
        \includegraphics[width=\textwidth, trim={0cm 2cm 6.5cm 2.4cm}, clip]{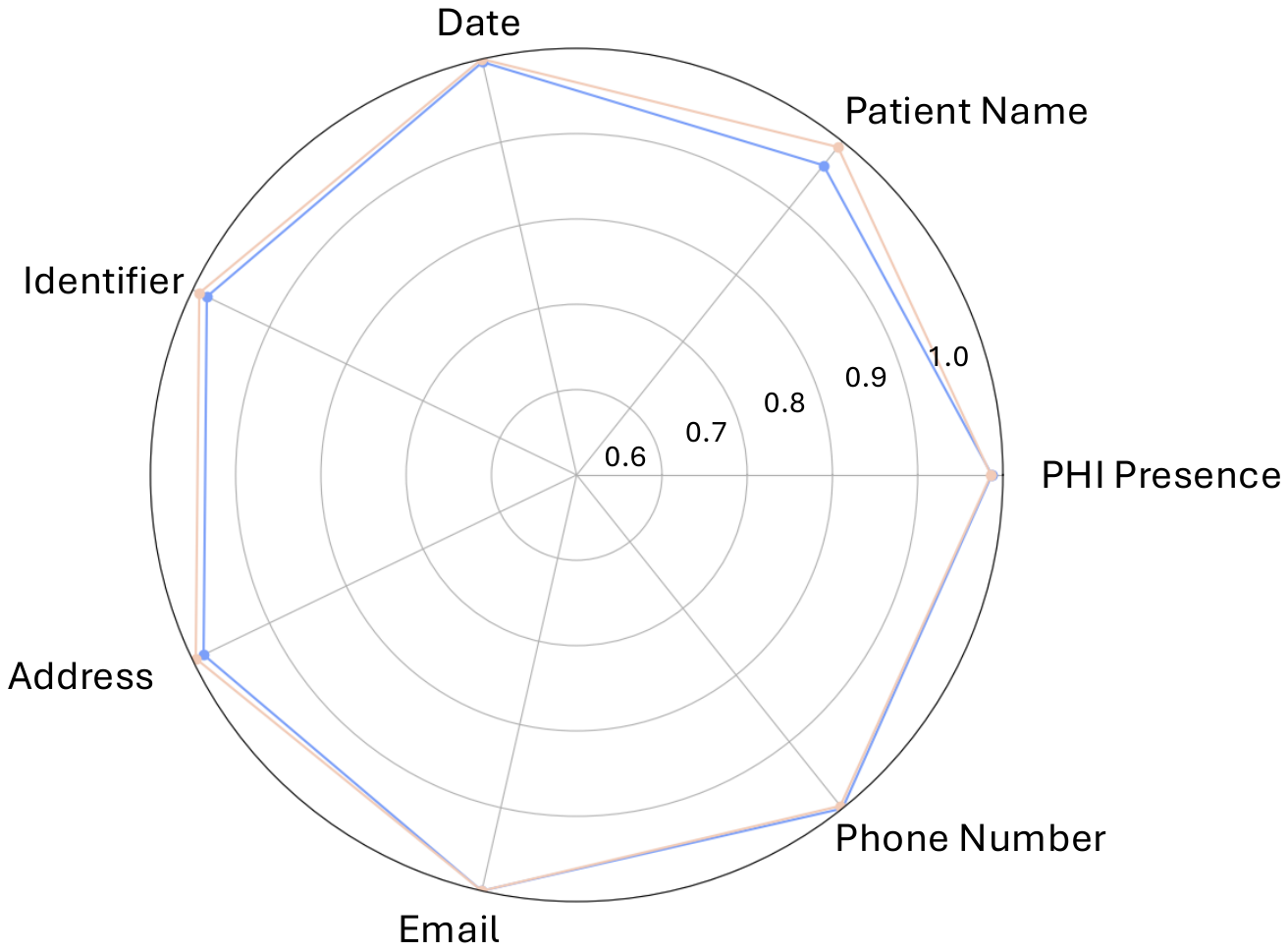}
        \caption{Instance-level Precision}
        \label{fig:instance-precision}
    \end{subfigure}  
    % \hfill
    \begin{subfigure}[b]{0.45\linewidth}
        \centering
        \includegraphics[width=\textwidth, trim={0cm 2cm 6.5cm 2.4cm}, clip]{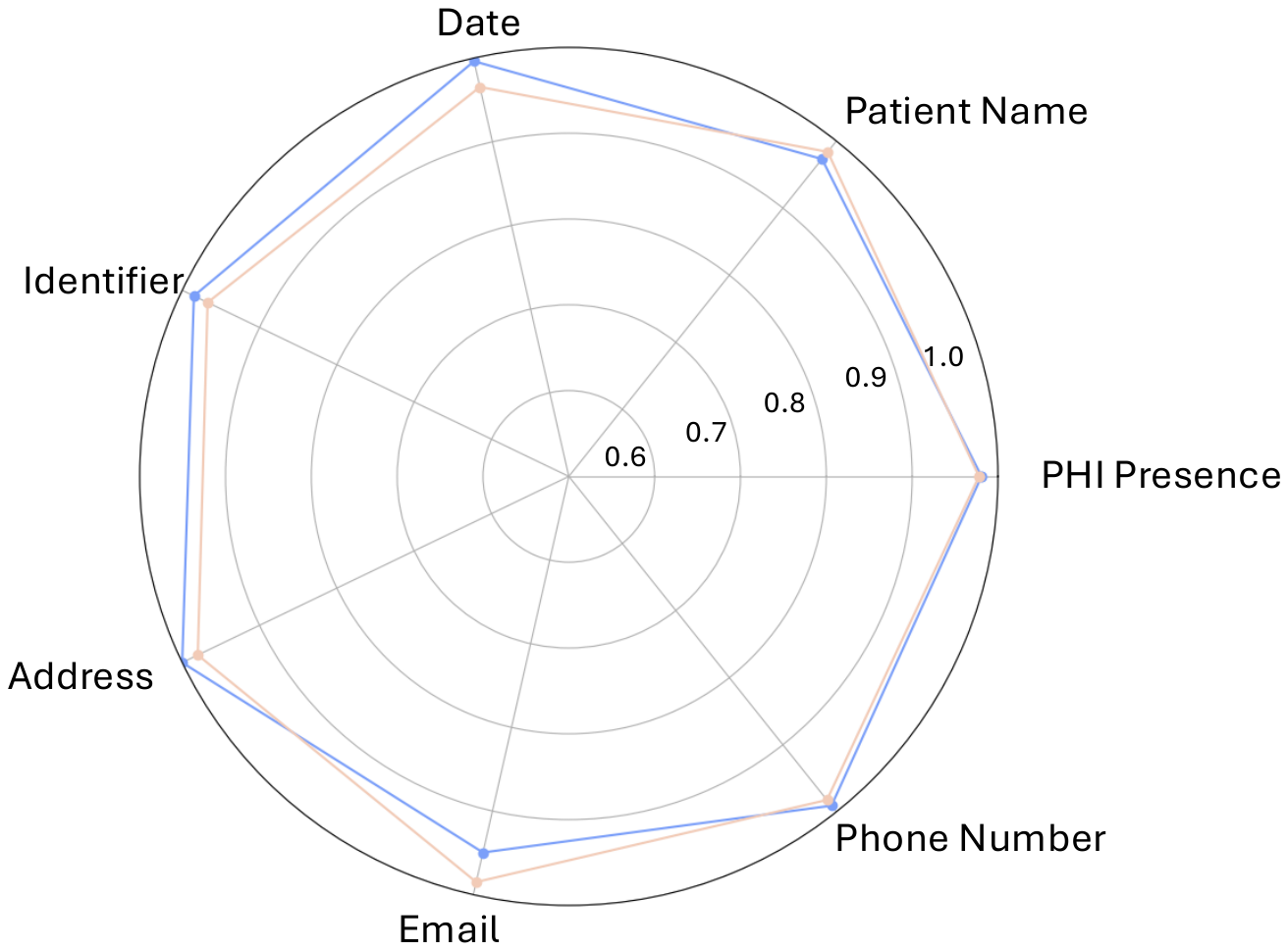}
        \caption{Instance-level Recall}
        \label{fig:instance-recall}
    \end{subfigure}
    \vskip\baselineskip
    \begin{subfigure}[b]{0.6\textwidth}
        \centering
    \includegraphics[width=\textwidth, trim={0cm 16cm 9cm 2cm}, clip]{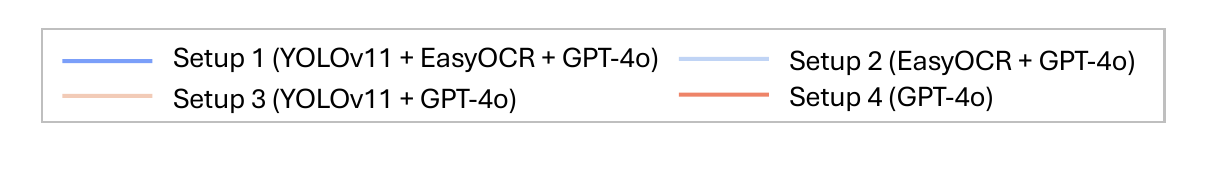}
        % \caption{Accuracy}
        \label{fig:legend}
    \end{subfigure}
    \caption{Comparison of precision and recall at \textbf{case and instance level}. Metric averages from five separate runs of each setup on the RadPHI-test dataset. Setups 2 and 4 are not evaluated at the instance level because of incompatible or absent predicted imprint coordinates. }
    \label{fig:all-metrics}
\end{figure}

\begin{table}[htp]
    \resizebox{\textwidth}{!}{%
    \centering
    \begin{tabular}{c c l l l l l l}
    \toprule
  & \textbf{No. Images}& \multicolumn{3}{c}{\textbf{Mean Precision [FP]}}                                   & \multicolumn{3}{c}{\textbf{Mean Recall [FN]}} \\       
  \cmidrule{3-8}
  & \textbf{with PHI} & \textbf{Worst Setup }& \textbf{Best Setup} & \textbf{Presidio} & \textbf{Worst Setup} & \textbf{Best Setup} & \textbf{Presidio }                          \\
  \hline
\cellcolor[HTML]{FFFFFF}PHI Presence  & 779& \cellcolor{setup2!25}0.9852 [11.4] & \cellcolor{setup1!25}1.0000 [0] & 0.9044 [54] & \cellcolor{setup4!25}0.9017 [76.6] & \cellcolor{setup3!25}0.9995 [0.4] & 0.6560 [268]\\
\cellcolor[HTML]{FFFFFF}Date & 267& \cellcolor{setup4!25}0.9309 [17.8] & \cellcolor{setup1!25}1.0000 [0] & 0.5734 [61] & \cellcolor{setup2!25}0.8689 [35] & \cellcolor{setup1!25}0.9963 [1] & 0.3071 [185] \\
\cellcolor[HTML]{FFFFFF}Patient  Name & 260& \cellcolor{setup4!25}0.7532 [72.2] & \cellcolor{setup3!25}1.0000 [0] & 0.4102 [220] & \cellcolor{setup4!25}0.8469 [39.8] & \cellcolor{setup3!25}0.9923 [2] & 0.5885 [107]\\
\cellcolor[HTML]{FFFFFF}Address & 248 & \cellcolor{setup4!25}0.9256 [17.8] & \cellcolor{setup3!25}0.9959 [1] & 0.7634 [31] & \cellcolor{setup4!25}0.8855 [28.4] & \cellcolor{setup1!25}1.0000 [0] & 0.4032 [148] \\
\cellcolor[HTML]{FFFFFF}Email         & 278 & \cellcolor{setup4!25}0.9950 [1.2] & \cellcolor{setup1!25}1.0000 [0] & 1.0000 [0]& \cellcolor{setup2!25}0.8489 [42] & \cellcolor{setup3!25}0.9849 [4.2] & 0.2086 [220] \\

\cellcolor[HTML]{FFFFFF}Phone Nr  & 260         & \cellcolor{setup4!25}0.9843 [3.2] & \cellcolor{setup1!25}1.0000 [0] & 0.9250 [6] & \cellcolor{setup4!25} 0.7631 [61.6] & \cellcolor{setup1!25}0.9908 [2.4] & 0.2846 [186]\\
\cellcolor[HTML]{FFFFFF}Identifier    & 237 & \cellcolor{setup2!25}0.8301 [44.2] & \cellcolor{setup3!25}1.0000 [0] & 0.4065 [146] & \cellcolor{setup4!25}0.8675 [31.4] & \cellcolor{setup1!25}0.9949 [1.2] & 0.4219 [137] \\
    \hline
    \end{tabular}
    }
    \caption{\hl{Worst and best performing setup based on mean precision and recall at \textbf{case level} computed across five different runs on RadPHI-test (1,000 images). The mean False-Positive (FP) and False-Negative (FN) values indicated in the brackets are computed for the corresponding setup across five different runs as well. Note that Presido operates deterministically, with the scores reported derived from a single run. The color-coding indicates the setup attaining the corresponding metric value, i.e.} \colorbox{setup1!25}{\strut} \hl{Setup 1}, \colorbox{setup2!25}{\strut} \hl{Setup 2}, \colorbox{setup3!25}{\strut} \hl{Setup 3}, \colorbox{setup4!25}{\strut} \hl{Setup 4.}}
    \label{tab:result-summary-case}
\end{table}

\begin{table}[htp]
    \centering
    \resizebox{\textwidth}{!}{%
    \begin{tabular}{c c l l l l l l}
    \toprule
  & \textbf{No. Instances} & \multicolumn{3}{c}{\textbf{Mean Precision [FP]}}                                   & \multicolumn{3}{c}{\textbf{Mean Recall [FN]}} \\       
  \cmidrule{3-8}      
  & \textbf{with PHI} & \textbf{Worst Setup }& \textbf{Best Setup} & \textbf{Presidio} & \textbf{Worst Setup} & \textbf{Best Setup} & \textbf{Presidio}                          \\
  \hline
\cellcolor[HTML]{FFFFFF}PHI Presence  & 1550 & \cellcolor{setup3!25}0.9859 [21.6] & \cellcolor{setup1!25}0.9869 [20.2] & 0.3002 [1074] & \cellcolor{setup3!25}0.9778 [34.4] & \cellcolor{setup1!25}0.9808 [29.8] & 0.3019 [1082]\\
\cellcolor[HTML]{FFFFFF}Date & 267 & \cellcolor{setup1!25}0.9963 [1] & \cellcolor{setup3!25}1.0000 [0] & 0.2985 [188] & \cellcolor{setup3!25}0.9655 [9.2] & \cellcolor{setup1!25}0.9963 [1] & 0.2809 [192] \\
\cellcolor[HTML]{FFFFFF}Patient  Name & 260 & \cellcolor{setup1!25}0.9642 [9.4] & \cellcolor{setup3!25}0.9915 [2.2] & 0.1511 [1117] & \cellcolor{setup1!25}0.9731 [7] & \cellcolor{setup3!25}0.9838 [4.2] & 0.3923 [158]\\
\cellcolor[HTML]{FFFFFF}Address       & 248 & \cellcolor{setup1!25}1.0000 [0] & \cellcolor{setup3!25}1.0000 [0] & 0.7619 [70] & \cellcolor{setup3!25}0.9798 [5] & \cellcolor{setup1!25}1.0000 [0] & 0.3750 [155] \\
\cellcolor[HTML]{FFFFFF}Email         & 278 &\cellcolor{setup1!25}1.0000 [0] & \cellcolor{setup3!25}1.0000 [0] & 0.9655 [2]& \cellcolor{setup1!25}0.9496 [14] & \cellcolor{setup3!25}0.9849 [4.2] & 0.2014 [222] \\
\cellcolor[HTML]{FFFFFF}Phone Nr           & 260 & \cellcolor{setup3!25}0.9961 [1] & \cellcolor{setup1!25}1.0000 [0] & 0.8235 [15] & \cellcolor{setup3!25}0.9823 [4.6] & \cellcolor{setup1!25}0.9908 [2.4] & 0.2692 [190]\\
\cellcolor[HTML]{FFFFFF}Identifier    & 237 & \cellcolor{setup1!25}0.9815 [4.4] & \cellcolor{setup3!25}0.9913 [2] & 0.2353 [312] & \cellcolor{setup3!25}0.9671 [7.8] & \cellcolor{setup1!25}0.9848 [3.6] & 0.3038 [165] \\
    \hline
    \end{tabular}
    }
    \caption{\hl{Worst and best performing setup based on mean precision and recall at \textbf{instance level} computed across five different runs on RadPHI-test (1,000 images). The mean False-Positive (FP) and False-Negative (FN) values indicated in the brackets are computed for the corresponding setup across five different runs as well. Note that Presido operates deterministically, with the scores reported derived from a single run. The color-coding indicates the setup attaining the corresponding metric value, i.e.} \colorbox{setup1!25}{\strut} \hl{Setup 1}, \colorbox{setup3!25}{\strut} \hl{Setup 3.}}
    \label{tab:result-summary-instance}
\end{table}

\begin{table}[htp]
    \centering
    \begin{tabular}{c l l l l l}
        \toprule
        \textbf{Setup} & \textbf{Models} & \textbf{Pipeline error rate} & \textbf{Total time} & \textbf{Prompt tokens} & \textbf{Response tokens} \\
        \midrule
        \cellcolor{setup1!25}1 & YOLOv11 + EasyOCR + GPT-4o & $0$ \%  & $4,143$ s & $1,727,290$ & $280,998$  \\
        \cellcolor{setup2!25}2 & EasyOCR + GPT-4o & $0$ \%  & $3,235$ s  & $1,744,125$  & $298,087$  \\
        \cellcolor{setup3!25}3 & YOLOv11 + GPT-4o & $1.6$ \%  & $7,685$ s  & $6,152,369$  & $474,746$  \\ 
        \cellcolor{setup4!25}4 & GPT-4o & $5.18$ \%  & $4,104$ s  & $3,530,302$ & $287,229$  \\
        \hline
        Baseline & Presidio & $0$ \% & $288$ s & N/A & N/A \\
         \bottomrule
         
    \end{tabular}
    \caption{\hl{Statistics related to pipeline execution: mean pipeline error (failure) rate, mean total time of analysis, the mean number of prompt and response tokens for 1,000 test cases across five runs for each setup evaluated on RadPHI-test. The pipeline error rate is the percentage of images that fail to process due to server issues or the presence of banned words in the requests sent to GPT-4o. The total time is estimated as the sum of the total processing and analysis time required for all 1,000 test cases.}}
    \label{tab:setups-stats}
\end{table}

\begin{figure}[hpt]
    \centering
    \includegraphics[width=\textwidth, trim={0cm 3cm 0cm 0cm}, clip]{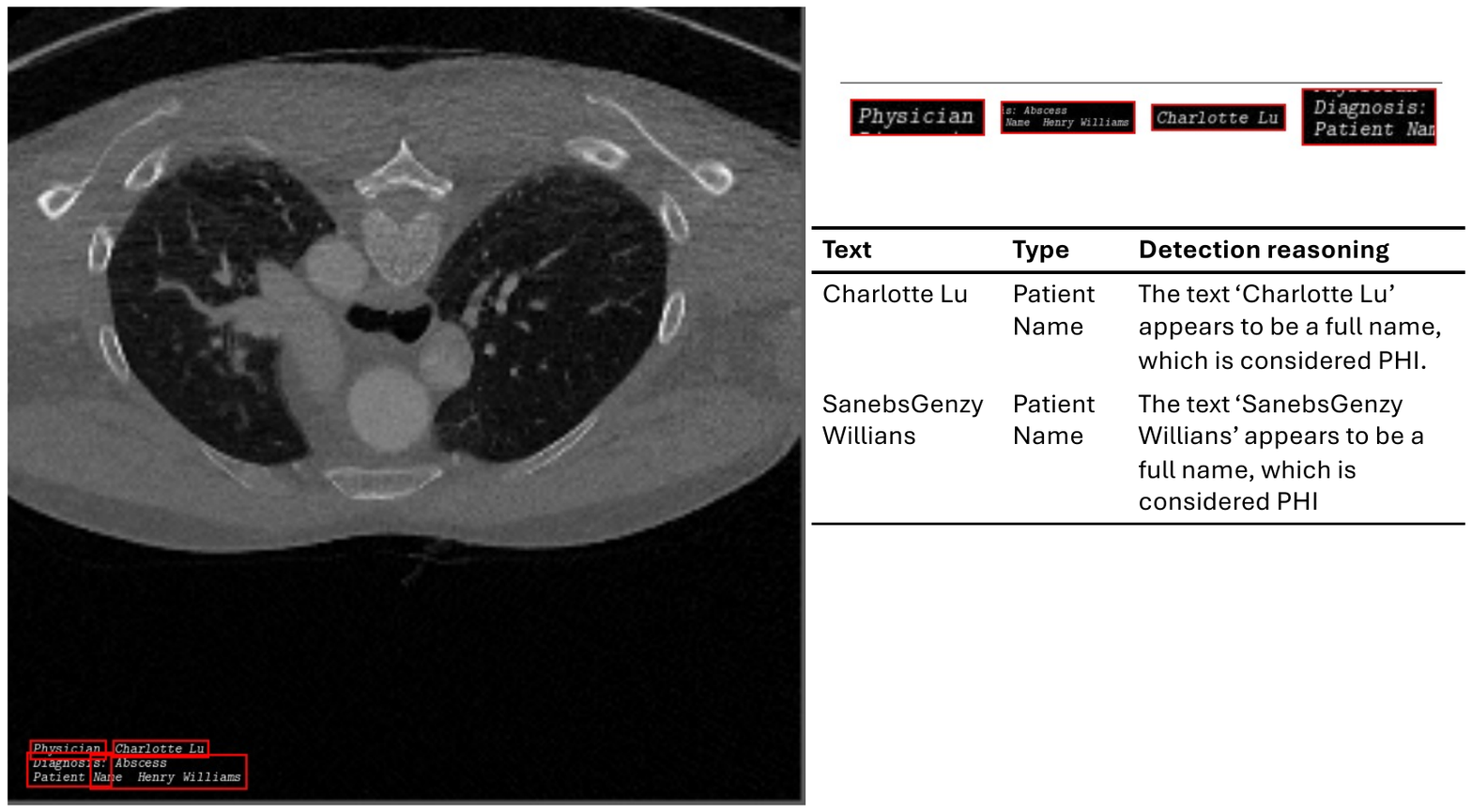}
    \caption{\hl{When text appears in multi-line blocks, out-of-the-box EasyOCR tries to fit bounding boxes that cover the entire text space. However, the bounding boxes might break the context consistency of the text. For example, Physician and Charlotte Lu are separated into two different crops, making it harder for the text analyzer to classify PHI content, particularly when a more fine-grain classification of names is required, e.g., physician or patient name.}}
    \label{fig:easyocr-failed-case}
\end{figure}

The results of the four setups are summarized in Figure~\ref{fig:all-metrics}, \hl{showing the precision and recall in classifying images for PHI and the corresponding six PHI categories at the case and instance level}. The worst and best setups based on precision and recall at two levels are summarized in Table~\ref{tab:result-summary-case}-\ref{tab:result-summary-instance}.
\paragraph{PHI presence detection}
\hl{Setups 1 and 3 demonstrate across all categories and metrics strong performance in detecting PHI presence at both the case and instance level. Setup 2, which relies on localization and text extraction by EasyOCR, exhibits a varying performance for different PHI categories. Setup 4, which utilizes GPT-4o end-to-end, obtains the lowest precision and recall. Setups 1 and 3 consistently achieve the highest results, attaining a maximum mean precision of 1.0 ± 0.0 (Setup 1) and a mean recall of 0.9995 ± 0.0007 (Setup 3) at the case level. Notably, Setup 3, with a case-level mean recall of 0.9995 (FN: 0.4 ± 0.55), indicates that, on average, fewer than one image containing at least one PHI is missed. However, with stricter evaluation at the instance level, the best precision and recall values decrease to 0.9869 ± 0.0008 and 0.9808 ± 0.0012 for Setup 1, resulting in an average of 29.8 ± 1.93 instances containing PHI being missed in the best-performing setup. Setups 2 and 4 are only assessed at the case level and show inferior performance compared to Setups 1 and 3. Additionally, Presidio falls behind all setups in both case and instance evaluations. At the case level, Presidio achieves precision and recall rates of 0.9044 and 0.6560, respectively, which drop significantly to 0.3002 and 0.3019 when evaluated at the instance level. This indicates that the tool has difficulty completely identifying all instances of PHI within the images.}

\paragraph{PHI category detection}
\hl{The performance of Setup 1 and Setup 3 is notably superior across all categories, while Setup 2 and Setup 4 significantly lag behind. Furthermore, Setups 1 and 3 demonstrate consistent performance across all metrics, in contrast to Setups 2 and 4, which exhibit considerable variability. Specifically, in terms of precision, Setups 2 and 4 achieve perfect scores for email and phone numbers but struggle with patient names and identifiers. Conversely, regarding recall, email and phone numbers experience the worst performance in Setup 2 and 4. Emails and phone numbers have unique formatting and signaling words, which allow easier differentiation from other types and hence higher precision. However, when texts are incompletely extracted or when key signaling elements are missed, it becomes challenging for the text analyzer to classify the content accurately. In these instances, email addresses and phone numbers can be easily confused with patient names and identifiers, as the former often include names while the latter consist of numbers. One of the reasons for incomplete text extraction is illustrated in Figure{~\ref{fig:easyocr-failed-case}}. EasyOCR in Setup 2 can break text continuity into separate bounding boxes, resulting in missing context and complicating the subsequent analysis of imprints. Similar to Setup 2 and Setup 4, Presidio also shows a high precision for selected PHI categories (phone numbers, email addresses) but low metric values on all other categories. In terms of recall, Presidio does not yield satisfactory results across all categories.}

\paragraph{Tokens, latency, and pipeline error rate}
Table~\ref{tab:setups-stats} compares the pipeline error rate, total time taken to process 1,000 images, and the number of prompt and response tokens used by GPT-4o across four setups. The pipeline error rate indicates the percentage of images that couldn't complete the process due to GPT-4o's inability to handle the request or the presence of \hl{blacklisted words in the query. The blacklisted words are often encountered by erroneous OCR outputs, especially noisy or misrecognized names, can inadvertently resemble words that trigger these safety mechanisms. However, these encounters are rare compared to the general request processing error.} While Setup 3 offers a slightly better overall performance in a few PHI categories, it has a significant drawback: the total analysis time is more than twice that of Setup 1. This increase is primarily due to the latency associated with querying GPT-4o multiple times for extracting text from individual image crops. On average, the number of prompt and response tokens in Setup 3 is three times and two times higher than that in Setup 1, as shown in Table~\ref{tab:setups-stats}. Using GPT-4o for the entire pipeline is technically feasible, but it does carry a higher rate of pipeline errors and hallucinations. These errors frequently arise from unstructured outputs or the generation of special tokens that can cause the model to fail to respond. \hl{In comparison to all four setups, Presidio has the shortest processing time (on average $0.288$s per case), which is approximately a factor 10 faster than the LLM-based pipeline from Setup 2 (on average $3.235$s per case). In regard to inference speed,  lightweight or rule-based models, which do not require significant computational resources, clearly outperform our GPT-4o-based setup. It is important to note that our PHI pipeline has not yet been optimized for throughput. Smaller models and optimized inference processes will allow us to significantly reduce computation time in the future.}

\subsection{MIDI}
 For further evaluation on the MIDI dataset, we choose Setup 1 over Setup 3 as the gain in performance of the latter is very limited whereas inference costs,  latency and token costs are much more favorable for Setup 1, see Table \ref{tab:setups-stats}.

In contrast to our simulated RadPHI-test dataset, the MIDI dataset offers a realistic benchmark which is more challenging due to its increased diversity and a larger number of imprints. Relying on signaling words might not be sufficient to detect PHI effectively. For example, an image series by itself may not be classified PHI, but if it includes a date element, as shown in Figure~\ref{fig:midi-data}, it is considered PHI. 

\begin{figure}
    \centering
    \includegraphics[width=\textwidth, trim={0cm 3cm 0cm 3cm}, clip]{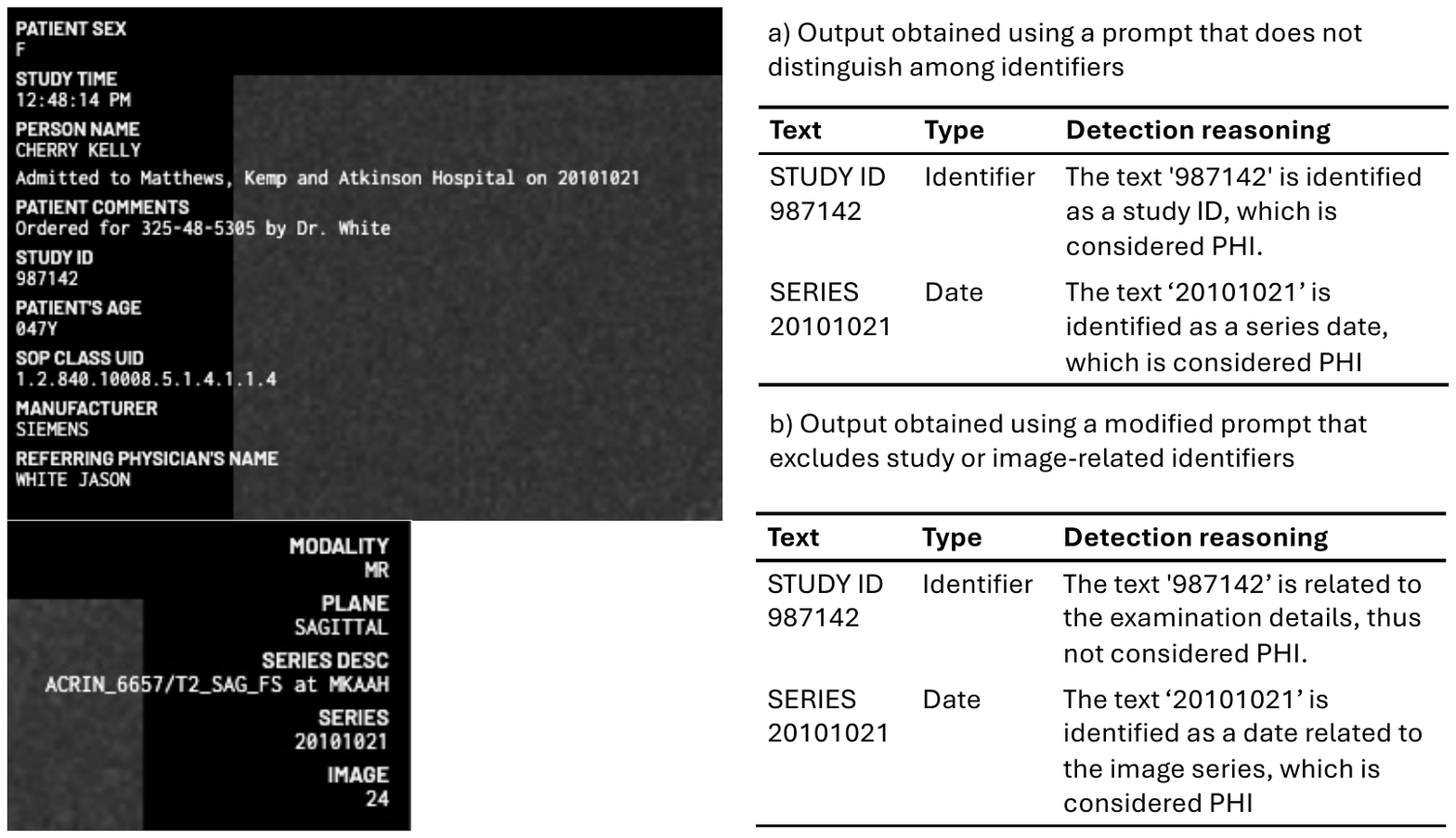}
    \caption{(Left) DICOM tag overlays with the DICOM viewer. (Right) Pipeline output for Setup 1 on the MIDI dataset: (a) using a prompt that considers all identifiers as PHI, and (b) with a modified prompt excluding study/image identifiers. Despite SERIES being non-PHI in the new prompt, the LLM can still detect the date element.}
    \label{fig:midi-result-example}
\end{figure}

\hl{When employing the same prompt used for RadPHI-test, we obtain the case-level precision and recall of 0.9509 ± 0.0127 and 0.9986 ±  0.0030, respectively as shown in Table~{\ref{tab:midi-results}} (a). On average, less than one image (FN: 0.2 ± 0.45) with PHI is overlooked. At the instance level where precise matching is crucial, precision and recall decline to 0.8797 ± 0.0096 and 0.9639 ± 0.0047, respectively. Key categories demonstrating strong performance include phone numbers, patient names, and dates. The precision of the identifier category is low due to misclassifications of numerous image- and study-related identifiers as PHI. The lower precision observed in the identifier category can be attributed to the misclassification of many image- and study-related identifiers as PHI. This issue stems from our initial guidelines for RadPHI-test, which regard all types of identifiers as PHI. To correct this, we refine the prompt by explicitly stating that study or image-related identifiers should not be categorized as PHI. The resulting enhancement in precision for the identifier category is reflected in Table~{\ref{tab:midi-results}} (b) and Figure~{\ref{fig:midi-result-example}}. The effects of the new prompt are evident not only in the improvement across all categories at the instance level but also in a minor reduction in performance at the case level for categories such as patient name and phone number. While this observation is intriguing, we do not delve deeply into further prompt optimization at this stage. However, it warrants investigation in our future work, where we plan to evaluate additional LLMs on a larger dataset and study this phenomenon. 

This prompt adaptation exemplifies how LLMs can be tailored to detect PHI in accordance with specific criteria, rather than adhering to a fixed definition. Ultimately, this flexibility allows users to configure the PHI detection system to better meet study-specific criteria.}

\begin{table}[htp]
    \centering
    \begin{tabular}{c c c l l l l}
        \toprule
        & &  \textbf{No. Images} / & \multicolumn{2}{c}{\textbf{(a) Same prompt as for RadPHI-test}} & \multicolumn{2}{c}{\textbf{(b) Modified prompt}} \\
        \cmidrule{4-7}
        \textbf{Level} & \textbf{Class} & \textbf{Instances} &  \textbf{Precision [FP]} & \textbf{Recall [FN]}  & \textbf{Precision [FP]} & \textbf{Recall [FN]}\\
        \midrule
        \multirow{6}{*}{Case} 
        & PHI & 147           & 0.9509 [7.6] & 0.9986 [0.2] & 0.9697 [4.6] \textsuperscript{(*)} & 1.0000 [0.0]\\
        & Date & 116          & 0.9931 [0.8] & 0.9914 [1.0] & 1.0000 [0.0] \textsuperscript{(*)} & 0.9914 [1.0]\\
        & Patient name & 104 & 0.9962 [0.4] & 0.9981 [0.2] & 1.0000 [0.0] & 0.9769 [2.4] \textsuperscript{(*)}\\
        & Address & 36       & 0.8931 [4.0] & 0.9222 [2.8] & 0.9500 [1.8] \textsuperscript{(*)} & 0.9444 [2.0]\\
        & Phone Nr  & 21    & 0.9193 [1.8] & 0.9714 [0.6] & 1.0000 [0.0] \textsuperscript{(*)} & 0.9143 [1.8] \textsuperscript{(*)}\\
        & Identifier & 79    & 0.7427 [26.8]& 0.9772 [1.8] & 0.9398 [5.0] \textsuperscript{(*)}& 0.9848 [1.2]\\
        \midrule
        \multirow{6}{*}{Instance} 
        & PHI    & 549       & 0.8797 [72.4] & 0.9639 [19.8] & 0.9721 [15.4] \textsuperscript{(*)} & 0.9781 [12] \textsuperscript{(*)} \\
        & Date & 207         & 0.9553 [8.8] & 0.9082 [19] & 0.9893 [2.2] \textsuperscript{(*)}& 0.9758 [5.0] \textsuperscript{(*)}\\
        & Patient name & 161  & 0.9851 [2.4]  & 0.9863 [2.2]  & 0.9776 [3.6]  & 0.9752 [4.9]\\
        & Address  & 39     & 0.8382 [7.0]  & 0.9282 [2.8]  & 0.9070 [3.8] \textsuperscript{(*)} & 0.9487 [2.0]\\
        & Phone Nr & 24     & 0.9073 [2.4]  & 0.9750 [0.6]  & 0.9648 [0.8] \textsuperscript{(*)}  & 0.9083 [2.2] \textsuperscript{(*)}\\
        & Identifier & 118   & 0.6589 [59.6] & 0.9746 [3.0]  & 0.9325 [8.4] \textsuperscript{(*)} & 0.9814 [2.2] \\
        \bottomrule
    \end{tabular}
    \caption{\hl{Performance comparison on the MIDI dataset using Setup 1: (a) performance with the RadPHI-test prompt and (b) performance with an adapted prompt for identifier classification. Superscripts (*) indicate statistical significance (pairwise Mann–Whitney U test, p < 0.05)}}
    \label{tab:midi-results}
\end{table}

% \begin{figure}
%     \centering
%     \begin{subfigure}[b]{0.49\linewidth}
%         \centering
%         \includegraphics[width=\textwidth, trim={0cm 2cm 2cm 2.7cm}, clip]{figures/MIDIResults.pdf}
%         \caption{Same system prompt as in RadPHI-test}
%     \end{subfigure}
%     \begin{subfigure}[b]{0.49\linewidth}
%         \centering
%         \includegraphics[width=\textwidth, trim={0cm 2cm 2cm 2.7cm}, clip]{figures/MIDIResultsAblation.pdf}
%         \caption{Modified system prompt for identifier}
%     \end{subfigure}
%     \caption{Performance comparison of precision, recall, F1 score, and accuracy on the MIDI dataset using Setup 1: (a) performance with the RadPHI-test prompt and (b) performance with an adapted prompt for identifier classification.}
%     \label{fig:midi-results}
% \end{figure}

\section{Discussion}
\paragraph{Text localization} For the localization step, we recommend using a dedicated object detection model instead of relying on the built-in localization component of a generic OCR model (Setup 2) or the localization capability of a multi-modal LLM (Setup 4). Our experimental results demonstrate that separating the localization step and using a localization model that can generalize well across different medical imaging modalities ensures strong overall performance, even when out-of-the-box OCR tools, like EasyOCR, are used downstream for text extraction (Setup 1). End-to-end GPT-4o (Setup 4) has the drawback of not returning the bounding box coordinates of text regions, which are essential for subsequent PHI redaction. Additionally, imprints are often incompletely identified and hallucinations occur much more frequently compared to other setups, resulting in lower performance. Moreover, in this setup, more pipeline errors related to server issues or unstructured output occur, which prevents GPT-4o from generating proper responses.

% For the text extraction step, both out-of-the-box OCR model (i.e., EasyOCR) and multi-modal LLM (i.e., GPT-4o) can yield strong performance.

\paragraph{Text extraction} When evaluating the OCR performance separately, GPT-4o outperforms EasyOCR by having lower word-level and character-level errors. However, for the end-to-end pipeline performance, the improvement in PHI detection when switching to GPT-4o from EasyOCR (i.e, Setup 3 vs. Setup 1) is minimal. This is because minor OCR errors produced by EasyOCR can easily be compensated by the robust language model in the text analysis step. While GPT-4o offers superior performance in text extraction, the overall gain in pipeline performance is modest and accompanied by higher latency and increased costs associated with token generation. Therefore, EasyOCR is preferred in our study. 

% 
% We attribute these slight improvements of GPT-4o over EasyOCR as the text analyzer, to the LLM's capabilities to effectively reason with minor errors in text extraction.
% 

\begin{figure}[htp]
    \centering
    \includegraphics[width=\textwidth, trim={0cm 4cm 0cm 2cm}, clip]{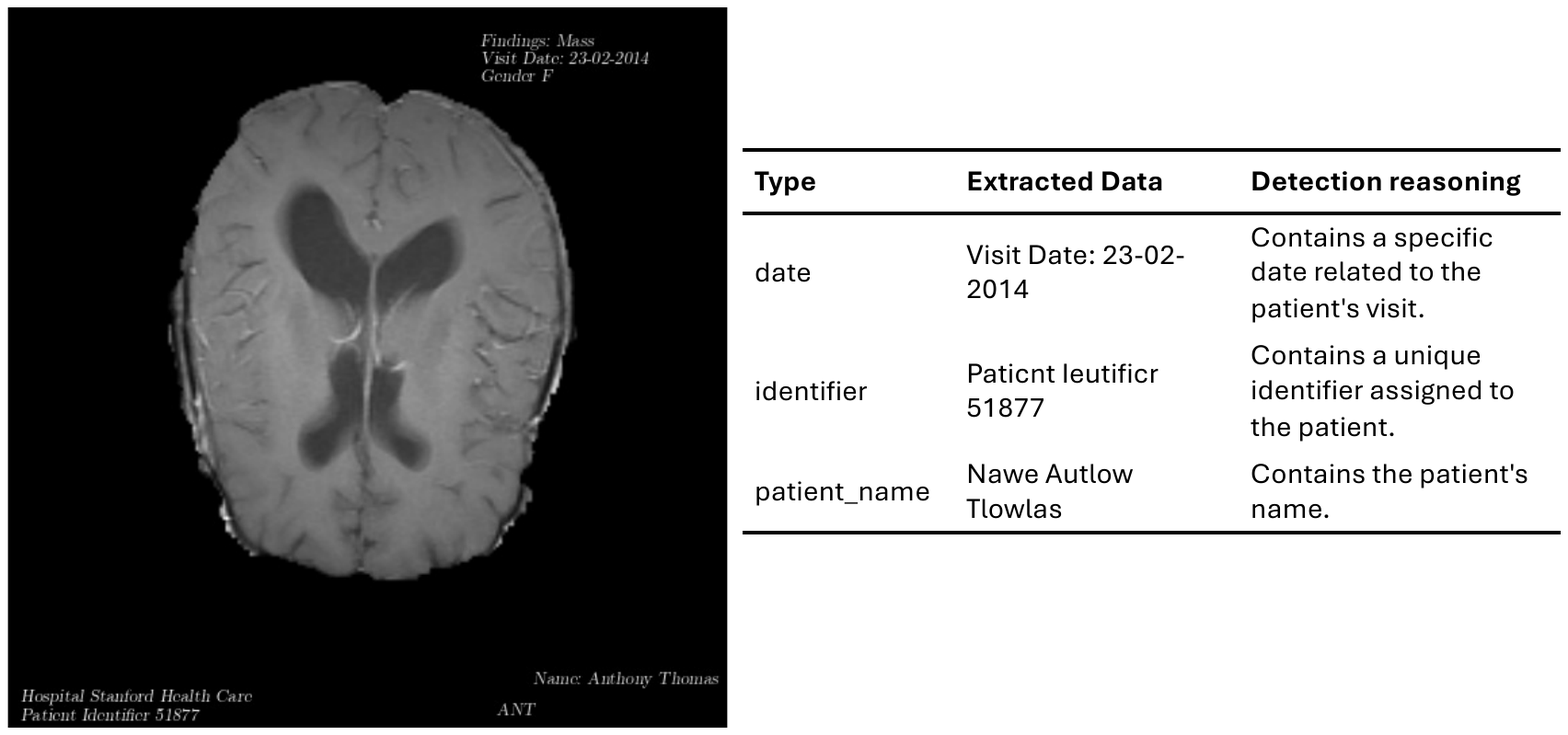}
    \caption{(Left) An image from the RadPHI-test dataset with simulated imprints. (Right) Results from the pipeline utilizing EasyOCR for text extraction and GPT-4o for text analysis. Despite the presence of spelling errors in the extracted text, GPT-4o effectively recognizes and interprets the content.}
    \label{fig:example-volume-309}
\end{figure}

\paragraph{Text analysis} For text analysis, GPT-4o is the default and only choice across all four different setups. The model excels at understanding text content and is particularly efficient in identifying PHI and further classifying it into sub-categories. Additionally, it can accurately discern context even when there are slight OCR errors in the extracted text. For instance, Figure~\ref{fig:example-volume-309} illustrates OCR errors related to dates, identifiers, and names. The model is capable of implicitly correcting these errors and reasoning based on the corrected text. Furthermore, an outstanding strength of LLM-based text analysis is the possibility to easily configure rules, sequential steps, and the output structure of the analysis through prompts. This adaptability makes it easy to redefine PHI and associated tasks across various studies or to accommodate different guidelines.

\paragraph{Generalization} \hl{A key concern for any AI-based system is its ability to generalize to unseen data. In context of PHI detection, we believe the main source of variability stems from the text representation such as font size, color, and background, as well as the content itself, which may include PHI in various languages or formats not explicitly defined within the default PHI criteria. We address this by leveraging pretrained models whenever possible, requiring minimal fine-tuning. By choosing pretrained models, we preserve the models' learned abstractions and only guide their reasoning towards specific criteria. For example, in our optimal setup (Setup 1), the only model that requires fine-tuning is the YOLOv11 model for imprint localization. By covering key imaging modalities, generating diverse imprints and utilizing numerous data augmentations methods during training, we achieve during inference that the YOLOv11 model performs robustly in unseen modalities that are not encountered during training, including PET, mammography, and ultrasound as part of the MIDI dataset. The other two models, namely EasyOCR and GPT-4o, work robustly with either no fine-tuning or only minor adjustments in the case of GPT-4o. On the MIDI dataset, which we have little control over the format of the imprints that appear in the images exported from the viewer. The results show that even without optimizing the prompt to tailor the formatting or representation, we achieve a decent recall rate of 0.9639, missing on average 19.8 out of 549 PHI instances. The precision is low, 0.8797, with 72.4 false positives on average due to unadjusted guidelines for classification. However, after adjusting the prompt to define specific criteria for inclusion and exclusion, we reduce the false positives to 15.4, improving the precision to 0.9721. This adaptation occurs only in the text analysis to conform to the new guidelines we select, while the text localization and extraction processes continue to perform well, even with the new dataset. While it is true that the LLM may struggle with new guidelines not covered in the initial prompt, modifying the PHI classification criteria is relatively straightforward using natural language. This stands in contrast with the complex engineering needed to customize rule-based classification systems or to fine-tune models with fixed rules. Additionally, it is less complicated to extend PHI classification to encompass diverse languages with LLMs, which is an advantage of using LLMs for this purpose.}

\paragraph{Capabilities of multi-modal LLM}The expansion of GPT-4o's capabilities beyond basic text analysis to the interpretation of small image crops and, ultimately, entire images reveals interesting findings. While the model demonstrates exceptional performance when analyzing image crops (Setup 3), it encounters a significant increase in technical errors and hallucinations when processing complete images in Setup 4 (Table~\ref{tab:setups-stats}). This inefficiency results in incomplete and inaccurate detection of PHI. It also underscores the current limitations of the multimodal language model, particularly in its capacity to handle images with diverse content and to perform complex reasoning tasks that require the integration of information from both images and text. Moreover, the model's non-deterministic behavior can lead to varying responses for identical inputs. However, by employing structured output formats with Pydantic \cite{pydantic} and setting the temperature parameter to zero, we have observed a marked improvement in response consistency, as indicated by the low standard deviation (Table~\ref{tab:setup1}-\ref{tab:setup4} in Supplementary Material). There remains significant potential to further explore the full capabilities of language models for the PHI detection task. In our future work, we plan to investigate and benchmark various prompting strategies and assess the model's robustness in languages other than English.  Additionally, given the high level of privacy required for medical data, it is essential to either fine-tune a LLM or host it in-house. Smaller LLM architectures are here of particular interest due to their ease of deployment, increased inference speed, reduced costs and the ability to customize them for specific tasks.

\paragraph{Limitations and future work} \hl{In our study, we focus on identifying the most effective setup for constructing a PHI pipeline, without delving deeply into benchmarking specific model types or variants to determine which offers the best performance. By optimizing the models used in each module separately, we acknowledge that our findings may not necessarily apply to particular models. For instance, we believe that some commercial OCR tools can process images effectively without requiring further fine-tuning, suggesting that Setup 2 may perform comparably to Setup 1. Moreover, our evaluation does not include testing our proposed approaches on real-world data containing PHI. In general, it is challenging to obtain medical image data containing actual PHI imprints for research and analysis in a privacy compliant way. Regulatory guidelines require that detected PHI is removed or de-identified by the data processor. In view of this, the MIDI dataset is created by utilizing an industry-standard medical image viewer to overlay synthetically generated PHI on medical images. Exporting image data directly from this viewer allows us to generate test data that is very close to reality.  We recognize that real-world datasets are very heterogeneous and may comprise a diverse range of modalities and scanners. Hence, we plan to further extend our benchmark in future work to incorporate a wider variety of model types and curate additional datasets for validation, and extend the language beyond English.}

\section{Conclusion}
In this paper, we dissect a pixel-level PHI detection pipeline into fundamental components which are implemented using well-established and generic foundation models.
In detail, we assess various pipeline setups utilizing YOLOv11 as an object detection model, EasyOCR as an OCR model, and GPT-4o as a text analysis model. Our findings indicate that employing LLM for the analysis phase yields remarkable contextual understanding and offers the flexibility to configuration adjustment. For the text localization and extraction phases, the optimal approach combines a fine-tuned object detection model with an out-of-the-box OCR tool. This combination outperforms pipelines relying on the OCR model's built-in text detection. Notably, we demonstrate that the LLM can extract text from image crops with performance levels that are on par with, or even exceed, those of established OCR tools like EasyOCR. However, while utilizing GPT-4o as an end-to-end PHI detection solution is a feasible approach, the setups relying on dedicated vision models designed for comprehensive image analysis are still superior in this case. This gap points to a promising area for future research aimed at enhancing vision-integrated LLM performance. 

% \bibliography{sample}

\begin{thebibliography}{99}
\bibitem{portability2012guidance}
U.S. Department of Health and Human Services: Guidance regarding methods for de-identification of protected health information in accordance with the health insurance portability and accountability act (HIPAA) privacy rule. \textit{Washington DC: U.S. Department of Health and Human Services}, 2012.

\bibitem{EuropeanParliament2016a}
European Parliament and Council of the European Union: Regulation (EU) 2016/679 of the European Parliament and of the Council of 27 April 2016 on the protection of natural persons with regard to the processing of personal data and on the free movement of such data, and repealing Directive 95/46/EC (General Data Protection Regulation). \textit{OJ L 119, 4.5.2016, p. 1--88}, 2016. Available at: https://data.europa.eu/eli/reg/2016/679/oj. Accessed 13 April 2023.

\bibitem{health2019cost}
Health Sector Cybersecurity Coordination Center: A Cost Analysis of Healthcare Sector Data Breaches. \textit{US Department of Health and Human Services, Washington, DC}, 2019. Available at: https://www.hhs.gov/sites/default/files/cost-analysis-of-healthcare-sector-data-breaches.pdf. Accessed 18 June 2025.

\bibitem{arpah_innovative_2024}
Advanced Research Projects Agency for Health (ARPA-H): Innovative Solutions Opening For Imaging Data Exchange (INDEX). \textit{Advanced Research Projects Agency for Health (ARPA-H)}, December 2024. Available at: https://sam.gov/opp/afb05a06fb1043ed99c8f39ab3021cbd/view. Accessed 9 April 2025.

\bibitem{kay_tesseract_2007}
Kay A: Tesseract: an open-source optical character recognition engine. \textit{Linux J} 2007(159):2, 2007.

\bibitem{kline_medical_2023}
Kline A, Appadurai V, Luo Y, Shah S: Medical image deidentification, cleaning and compression using Pylogik. \textit{arXiv}, DOI: 10.48550/arXiv.2304.12322, May 2023. Available at: http://arxiv.org/abs/2304.12322.

\bibitem{langlois_open_2024}
Langlois Q, Szelagowski N, Vanderdonckt J, Jodogne S: Open platform for the de-identification of burned-in texts in medical images using deep learning. In: Proceedings of the 17th International Joint Conference on Biomedical Engineering Systems and Technologies, Rome, Italy, pp. 297–304. \textit{SCITEPRESS - Science and Technology Publications}, 2024. DOI: 10.5220/0012430300003657. Available at: https://www.scitepress.org/DigitalLibrary/Link.aspx?doi=10.5220/0012430300003657. 

\bibitem{rempe_-identification_2024}
Rempe M, Heine L, Seibold C, Hörst F, Kleesiek J: De-identification of medical imaging data: A comprehensive tool for ensuring patient privacy. \textit{arXiv}, DOI: 10.48550/arXiv.2410.12402, October 2024. Available at: http://arxiv.org/abs/2410.12402.

\bibitem{macdonald_method_2024}
Macdonald JA, Morgan KR, Konkel B, Abdullah K, Martin M, Ennis C, Lo JY, Stroo M, Snyder DC, Bashir MR: A method for efficient de‑identification of DICOM metadata and burned‑in pixel text. \textit{J Imaging Inform Med}, DOI: 10.1007/s10278-024-01098-7, October 2024.

\bibitem{monteiro_-identification_2017} Monteiro E, Costa C, Oliveira JL: A de‑identification pipeline for ultrasound medical images in DICOM format. \textit{J Med Syst}, DOI: 10.1007/s10916-017-0736-1, April 2017.

\bibitem{vcelak_identification_2019} Vcelak P, Kryl M, Kratochvil M, Kleckova J: Identification and classification of DICOM files with burned-in text content. \textit{Int J Med Inform}, DOI: 10.1016/j.ijmedinf.2019.02.011, June 2019.

\bibitem{mdai} MD.ai. Available at https://docs.md.ai/deid/deid/. Accessed 28 April 2025.

\bibitem{gcloud_healthcare} Google Cloud. Available at https://cloud.google.com/healthcare-api/docs/concepts/de-identification. Accessed 9 December 2024.

\bibitem{johnsnowlabs} John Snow Labs. Available at https://www.johnsnowlabs.com/deidentification/. Accessed 9 December 2024.

\bibitem{noauthor_glendor_nodate} Glendor. Available at https://glendor.com/. Accessed 24 January 2025.

\bibitem{microsoft_presidio} Microsoft. Presidio, version 2.2.358. Available at https://microsoft.github.io/presidio/. Accessed June 2025.

\bibitem{jahan_comprehensive_2024} Jahan I, Laskar MT, Peng C, Huang JX: A comprehensive evaluation of large language models on benchmark biomedical text processing tasks. \textit{Comput Biol Med}, DOI: 10.1016/j.compbiomed.2024.108189, March 2024.

\bibitem{clunie_summary_2024} Clunie D, Prior F, Rutherford M, Moore S, Parker W, Kondylakis H, Ludwigs C, Klenk J, Lou B, O'Sullivan LT, Marcus D, Dobes J, Gutman A, Farahani K: Summary of the National Cancer Institute 2023 Virtual Workshop on Medical Image De-identification-Part 1: Report of the MIDI Task Group - Best Practices and Recommendations, Tools for Conventional Approaches to De-identification, International Approaches to De-identification, and Industry Panel on Image De-identification. \textit{J Imaging Inform Med}, DOI: 10.1007/s10278-024-01182-y, July 2024.

\bibitem{clunie_summary_2024-1} Clunie D, Taylor A, Bisson T, Gutman D, Xiao Y, Schwarz CG, Greve D, Gichoya J, Shih G, Kline A, Kopchick B, Farahani K: Summary of the National Cancer Institute 2023 Virtual Workshop on Medical Image De-identification-Part 2: Pathology Whole Slide Image De-identification, De-facing, the Role of AI in Image De-identification, and the NCI MIDI Datasets and Pipeline. \textit{J Imaging Inform Med}, DOI: 10.1007/s10278-024-01183-x, July 2024.

\bibitem{openai_gpt4o_nodate} OpenAI. Available at https://openai.com/index/hello-gpt-4o/. Accessed 20 December 2024.

\bibitem{wasserthal2023totalsegmentator} Wasserthal J, Breit HC, Meyer MT, Pradella M, Hinck D, Sauter AW, Heye T, Boll DT, Cyriac J, Yang S, et al.: TotalSegmentator: robust segmentation of 104 anatomic structures in CT images. \textit{Radiol Artif Intell} 5(5), 2023.

\bibitem{huang2022bs} Huang Z, Pu X, Tang G, Ping M, Jiang G, Wang M, Wei X, Ren Y: BS-80K: The first large open-access dataset of bone scan images. \textit{Comput Biol Med} 151:106221, 2022.

\bibitem{wang2017chestx} Wang X, Peng Y, Lu L, Lu Z, Bagheri M, Summers RM: ChestX-ray8: Hospital-scale chest x-ray database and benchmarks on weakly-supervised classification and localization of common thorax diseases. In: Proceedings of the IEEE conference on computer vision and pattern recognition, pp. 2097–2106, 2017.

\bibitem{antonelli2022medical} Antonelli M, Reinke A, Bakas S, Farahani K, Kopp-Schneider A, Landman BA, Litjens G, Menze B, Ronneberger O, Summers RM, et al.: The medical segmentation decathlon. \textit{Nat Commun} 13(1):4128, 2022.

\bibitem{infosagebaseorg_midi-b_nodate} Farahani K, Clunie D, Klenk J, Kopchick B, Diaz M, Pan Q, Pei L, Prior F, Rutherford M, Singh A, Sutton G, Wagner U: Medical Image De-Identification Benchmark (MIDI-B). Available at https://www.synapse.org/Synapse:syn53065760. Accessed 16 April 2025.

\bibitem{mdai_viewer} MD.ai. Available at https://www.md.ai. Accessed 28 April 2025.

\bibitem{ultralytics_yolo11_nodate} Ultralytics. Available at https://docs.ultralytics.com/models/yolo11. Accessed 5 December 2024.

\bibitem{redmon_you_2016} Redmon J, Divvala S, Girshick R, Farhadi A: You Only Look Once: Unified, Real-Time Object Detection. In: Proceedings of the 2016 IEEE Conference on Computer Vision and Pattern Recognition (CVPR), Las Vegas, NV, USA, pp. 779–788, DOI: 10.1109/CVPR.2016.91, June 2016.

\bibitem{noauthor_jaidedaieasyocr_nodate} EasyOCR. Available at https://github.com/JaidedAI/EasyOCR. Accessed 20 December 2024.

\bibitem{baek_character_2019} Baek Y, Lee B, Han D, Yun S, Lee H: Character region awareness for text detection. \textit{arXiv}, DOI: 10.48550/arXiv.1904.01941, April 2019.

\bibitem{baek2019STRcomparisons} Baek J, Kim G, Lee J, Park S, Han D, Yun S, Oh SJ, Lee H: What is wrong with scene text recognition model comparisons? Dataset and model analysis. In: Proceedings of the International Conference on Computer Vision (ICCV), 2019.

\bibitem{brown2020language} Brown T, Mann B, Ryder N, Subbiah M, Kaplan JD, Dhariwal P, Neelakantan A, Shyam P, Sastry G, Askell A, et al.: Language models are few-shot learners. \textit{Adv Neural Inf Process Syst} 33:1877–1901, 2020.

\bibitem{pydantic} Colvin S: Pydantic: Data validation and settings management using Python type annotations, version 2.10.6. Available at https://github.com/pydantic/pydantic. Accessed 28 April 2025.

\end{thebibliography}

\pagebreak
\appendix

\section{Results on RadPHI-test}
\renewcommand{\thetable}{A\arabic{table}}
\setcounter{table}{0}
\renewcommand{\thefigure}{A\arabic{figure}}
\setcounter{figure}{0}

\begin{table}[htp]
    \centering
    \begin{tabular}{c c c c c}
        \toprule
         \textbf{Class} & \textbf{Case Precision} & \textbf{Case Recall} & \textbf{Instance Precision} & \textbf{Instance Recall}  \\
         \midrule
PHI Presence   & 1.0000 ± 0.0000 & 0.9923 ± 0.0000 & 0.9869 ± 0.0008 & 0.9808 ± 0.0012 \\ \hline
Date         & 1.0000 ± 0.0000 & 0.9963 ± 0.0000 & 0.9963 ± 0.0000 & 0.9963 ± 0.0000 \\
Patient Name & 0.9778 ± 0.0032 & 0.9815 ± 0.0017 & 0.9642 ± 0.0033 & 0.9731 ± 0.0000\\
Address      & 0.9952 ± 0.0018 & 1.0000 ± 0.0000 & 0.9857 ± 0.0045 & 1.0000 ± 0.0000\\
Email        & 1.0000 ± 0.0000 & 0.9496 ± 0.0000 & 1.0000 ± 0.0000 & 0.9496 ± 0.0000\\
Phone Nr     & 1.0000 ± 0.0000 & 0.9908 ± 0.0021 & 1.0000 ± 0.0000 & 0.9908 ± 0.0021\\
Identifier   & 0.9916 ± 0.0000 & 0.9949 ± 0.0035 & 0.9815 ± 0.0023 & 0.9848 ± 0.0023\\
         \bottomrule
         
    \end{tabular}
    \caption{\hl{PHI detection performance of setup 1 (YOLOv11 + EasyOCR + GPT-4o) on RadPHI-test.}}
    \label{tab:setup1}
\end{table}

\begin{table}[htp]
    \centering
    \begin{tabular}{c c c}
        \toprule
         \textbf{Class} & \textbf{Case Precision} & \textbf{Case Recall} \\
         \midrule
PHI Presence   & 0.9852 ± 0.0007 & 0.9743 ± 0.0016\\ \hline
Date         & 0.9983 ± 0.0023 & 0.8689 ± 0.0065\\
Patient Name & 0.8547 ± 0.0011 & 0.9323 ± 0.0058\\
Address      & 0.9942 ± 0.0022 & 0.9750 ± 0.0018\\
Email        & 1.0000 ± 0.0000 & 0.8489 ± 0.0025\\
Phone Nr     & 0.9908 ± 0.0000 & 0.8254 ± 0.0021\\
Identifier   & 0.8301 ± 0.0027 & 0.9114 ± 0.0000\\
         \bottomrule
         
    \end{tabular}
    \caption{\hl{PHI detection performance of setup 2 (EasyOCR + GPT-4o) on RadPHI-test.}}
    \label{tab:setup2}
\end{table}

\begin{table}[htp]
    \centering
    \begin{tabular}{c c c c c}
        \toprule
         \textbf{Class} & \textbf{Case Precision} & \textbf{Case Recall} & \textbf{Instance Precision} & \textbf{Instance Recall}  \\
         \midrule
PHI Presence   & 0.9954 ± 0.0007  & 0.9995 ± 0.0007 & 0.9859 ± 0.0006 & 0.9778 ± 0.0021\\ \hline
Date         & 1.0000 ± 0.0000 & 0.9655 ± 0.0041 & 1.0000 ± 0.0000 & 0.9655 ± 0.0041\\
Patient Name &  1.0000 ± 0.0000 & 0.9923 ± 0.0047 & 0.9915 ± 0.0017 & 0.9838 ± 0.0032\\
Address      &  0.9959 ± 0.0029 & 0.9798 ± 0.0040 & 0.9959 ± 0.0029 & 0.9798 ± 0.0040\\
Email        &  1.0000 ± 0.0000 & 0.9849 ± 0.0053 & 1.0000 ± 0.0000 & 0.9849 ± 0.0053\\
Phone Nr     &  0.9961 ± 0.0000 & 0.9823 ± 0.0034 & 0.9961 ± 0.0000 & 0.9823 ± 0.0034\\
Identifier   &  1.0000 ± 0.0000 & 0.9755 ± 0.0046 & 0.9913 ± 0.0000 & 0.9671 ± 0.0046\\
         \bottomrule
         
    \end{tabular}
    \caption{\hl{PHI detection performance of setup 3 (YOLOv11 + GPT-4o) on RadPHI-test.}}
    \label{tab:setup3}
\end{table}

\begin{table}[htp]
    \centering
    \begin{tabular}{c c c }
        \toprule
         \textbf{Class} & \textbf{Case Precision} & \textbf{Case Recall}  \\
         \midrule
PHI Presence   & 0.9918 ± 0.0023  & 0.9017 ± 0.0066\\
\midrule
Date         & 0.9309 ± 0.0087 & 0.8966 ± 0.0146\\
Patient Name & 0.7532 ± 0.0084 & 0.8469 ± 0.0202\\
Address      & 0.9256 ± 0.0184 & 0.8855 ± 0.0175\\
Email        & 0.9950 ± 0.0054 & 0.8633 ± 0.0171\\
Phone Nr     & 0.9843 ± 0.0090 & 0.7631 ± 0.0169\\
Identifier   & 0.8809 ± 0.0230 & 0.8675 ± 0.0238\\
         \bottomrule
         
    \end{tabular}
    \caption{\hl{PHI detection performance of setup 4 (GPT-4o) on RadPHI-test.}}
    \label{tab:setup4}
\end{table}

\clearpage
\section{Results on MIDI}
\renewcommand{\thetable}{B\arabic{table}}
\setcounter{table}{0}
\renewcommand{\thefigure}{B\arabic{figure}}
\setcounter{figure}{0}
\begin{table}[htp]
    \centering
    \begin{tabular}{c c c c c}
        \toprule
         \textbf{Class} & \textbf{Case Precision} & \textbf{Case Recall}  & \textbf{Instance Precision} & \textbf{Instance Recall}\\
         \midrule
PHI Presence   & 0.9509 ± 0.0127 & 0.9986 ± 0.0030 & 0.8797 ± 0.0096 & 0.9639 ± 0.0047\\
\midrule
Date         &  0.9931 ± 0.0039 & 0.9914 ± 0.0000 & 0.9553 ± 0.0081 & 0.9082 ± 0.0048\\
Patient Name & 0.9962 ± 0.0052 & 0.9981 ± 0.0043 & 0.9851 ± 0.0055 & 0.9863 ± 0.0081\\
Address      &  0.8931 ± 0.0305 & 0.9222 ± 0.0232 & 0.9851 ± 0.0055 & 0.9863 ± 0.0081\\
Phone Nr     & 0.9193 ± 0.0186 & 0.9714 ± 0.0261 & 0.9073 ± 0.0189 & 0.9750 ± 0.0228\\
Identifier   & 0.7427 ± 0.0188 & 0.9772 ± 0.0057 & 0.6589 ± 0.0167 & 0.9746 ± 0.0060\\
         \bottomrule
         
    \end{tabular}
    \caption{\hl{PHI detection performance of setup 1 (YOLOv11 + EasyOCR + GPT-4o) on MIDI using the same prompt as RadPHI-test}}
    \label{tab:midi-unmodified-prompt}
\end{table}

\begin{table}[htp]
    \centering
    \begin{tabular}{c c c c c}
        \toprule
         \textbf{Class} & \textbf{Case Precision} & \textbf{Case Recall}  & \textbf{Instance Precision} & \textbf{Instance Recall}\\
         \midrule
PHI Presence   &  0.9697 ± 0.0035 & 1.0000 ± 0.0000 & 0.9721 ± 0.0032  & 0.9781 ± 0.0043\\
\midrule
Date         &  1.0000 ± 0.0000 & 0.9914 ± 0.0000 & 0.9893 ± 0.0064 & 0.9758 ± 0.0076\\
Patient Name &  1.0000 ± 0.0000  & 0.9769 ± 0.0129 & 0.9776 ± 0.0035 & 0.9752 ± 0.0116\\
Address      &  0.9500 ± 0.0116 & 0.9444 ± 0.0278 & 0.9070 ± 0.0092 & 0.9487 ± 0.0256\\
Phone Nr     & 1.0000 ± 0.0000 & 0.9143 ± 0.0213 & 0.9648 ± 0.0197  & 0.9083 ± 0.0186\\
Identifier   & 0.9398 ± 0.0136 & 0.9848 ± 0.0057 & 0.9325 ± 0.0115 & 0.9814 ± 0.0038\\
         \bottomrule
         
    \end{tabular}
    \caption{\hl{PHI detection performance of setup 1 (YOLOv11 + EasyOCR + GPT-4o) on MIDI using the modified prompt with focus on eliminating false positives in the identifier category.}}
    \label{tab:midi-modified-prompt}
\end{table}
\clearpage
\section{Training data for YOLOv11}
\renewcommand{\thetable}{C\arabic{table}}
\setcounter{table}{0}
\renewcommand{\thefigure}{C\arabic{figure}}
\setcounter{figure}{0}
\begin{figure}[htp]
    \centering
    \includegraphics[width=0.9\textwidth, trim={1cm 4.2cm 0cm 3cm}, clip]{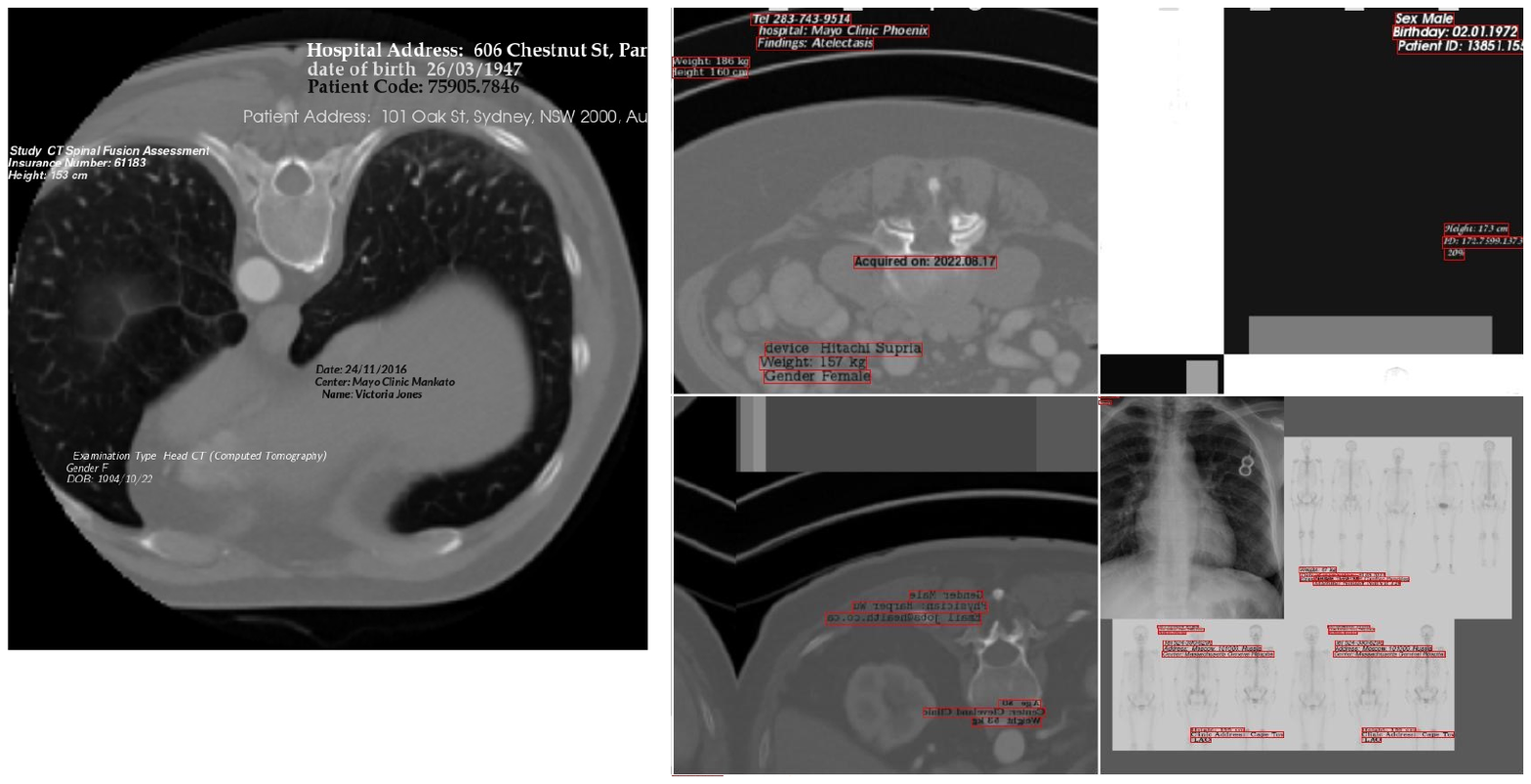}
    \caption{Examples of data used for training the localization model YOLOv11. \textbf{(Left)} Training image with multiple simulated text imprints in random locations \textbf{(Right)} Augmentation of training images. One of the augmentation schemes we use is mosaic augmentation, where the new image is created by blending multiple patches of original images. }
    \label{fig:yolo-training-data}
\end{figure}
\clearpage
\section{Examples of pipeline output}
\renewcommand{\thetable}{D\arabic{table}}
\setcounter{table}{0}
\renewcommand{\thefigure}{D\arabic{figure}}
\setcounter{figure}{0}

\begin{figure}[htp]
    \centering
    \includegraphics[width=\textwidth, trim={0cm 3cm 0cm 0cm}, clip]{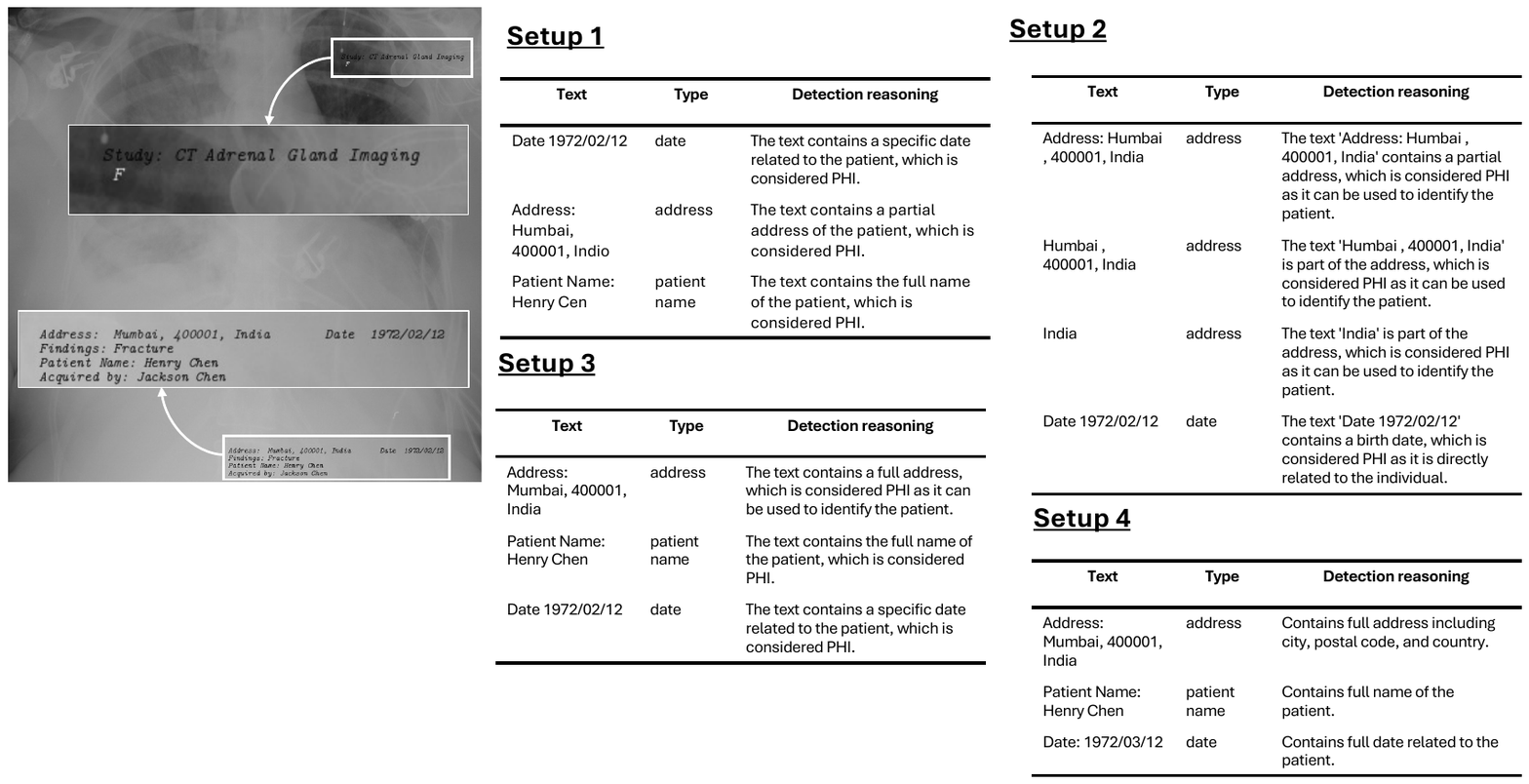}
    \caption{\hl{Comparison of outputs from four setups: Setups 1, 3, and 4 demonstrate similar response patterns. In contrast, Setup 2 exhibits duplicate text extraction, resulting in similar entries.}}
    \label{fig:example-volume-456}
\end{figure}

\begin{figure}[hpt]
    \centering
    \includegraphics[width=\textwidth, trim={0cm 8cm 0cm 0cm}, clip]{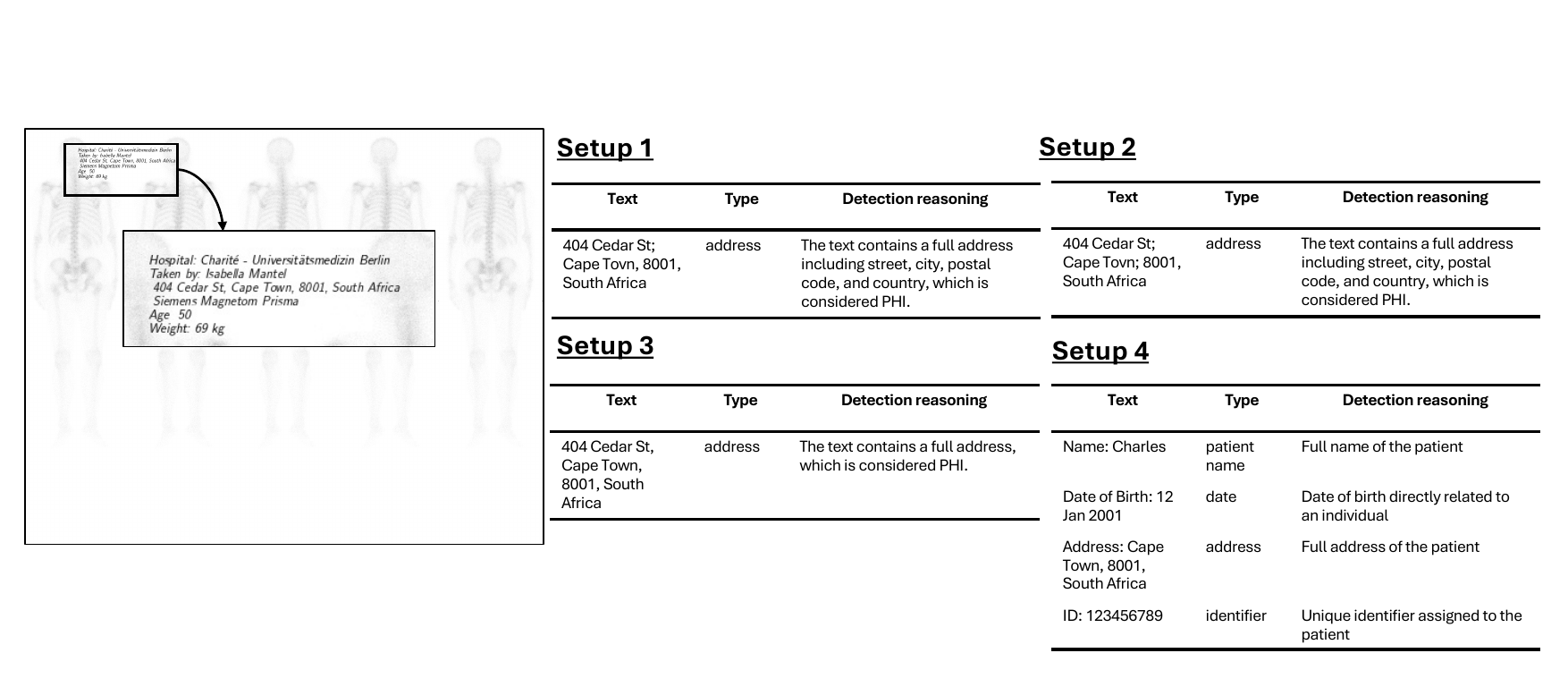}
    \caption{\hl{Comparison of outputs from four setups: Setup 1, 2, and 3 accurately classify PHI category while Setup 4 illustrates model hallucination, generating responses that are not based on the provided input.}}
    \label{fig:example-volume-11}
\end{figure}

\begin{figure}[hpt]
    \centering
    \includegraphics[width=1\textwidth, trim={0cm 6cm 0cm 0cm}, clip]{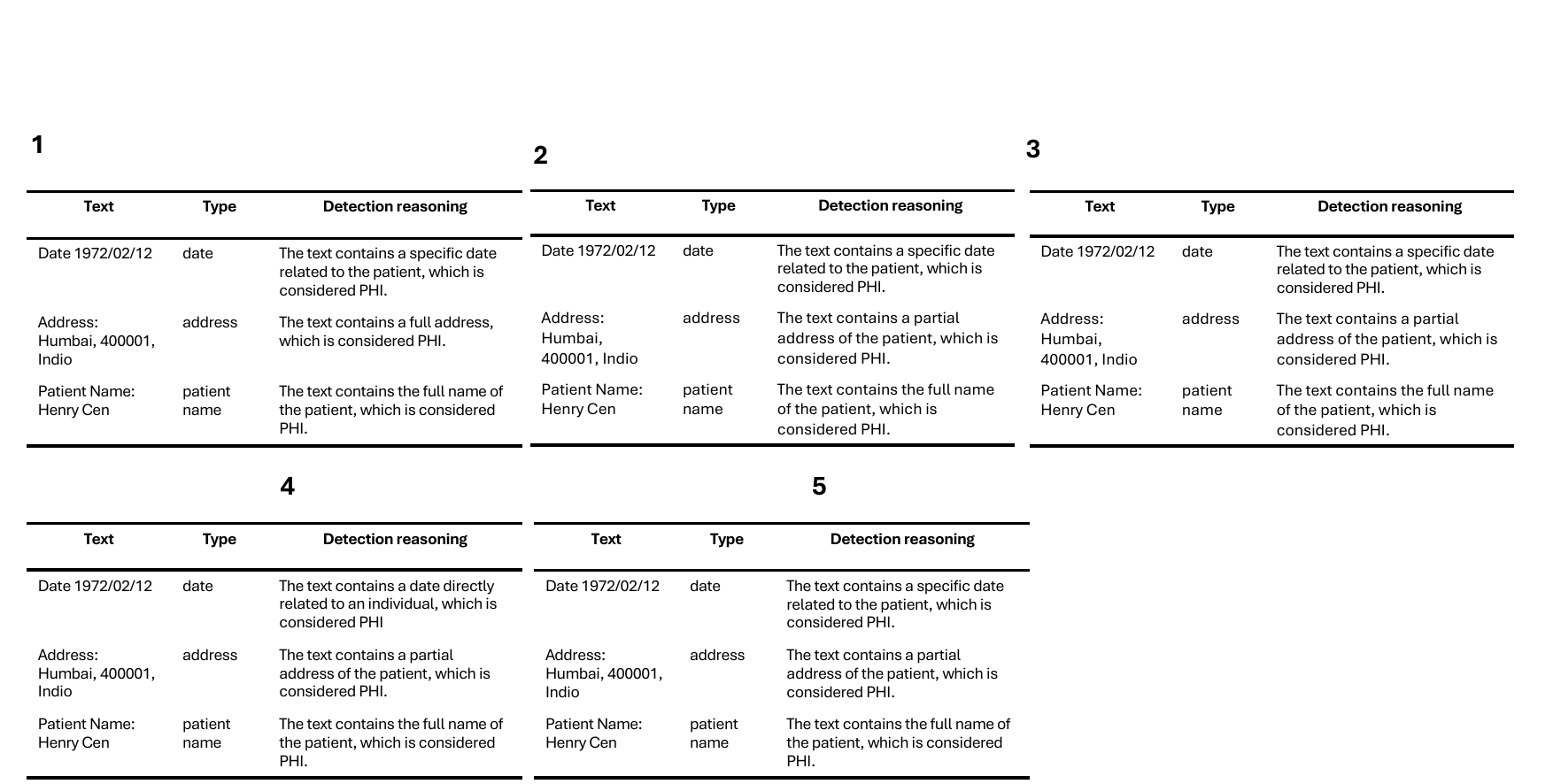}
    \caption{\hl{Comparison of responses from the same case across five distinct runs. Utilizing Pydantic and a temperature setting of 0 enhances the consistency of responses, even though some variation in reasoning is observed across the runs.}}
    \label{fig:example-consistency}
\end{figure}

% Acknowledgements should be brief, and should not include thanks to anonymous referees and editors, or effusive comments. Grant or contribution numbers may be acknowledged.

% \section*{Author contributions statement}

% Must include all authors, identified by initials, for example:
% A.A. conceived the experiment(s),  A.A. and B.A. conducted the experiment(s), C.A. and D.A. analysed the results.  All authors reviewed the manuscript. 

% \section*{Additional information}

% To include, in this order: \textbf{Accession codes} (where applicable); \textbf{Competing interests} (mandatory statement). 

% The corresponding author is responsible for submitting a \href{http://www.nature.com/srep/policies/index.html#competing}{competing interests statement} on behalf of all authors of the paper. This statement must be included in the submitted article file.

% \begin{figure}[ht]
% \centering
% \includegraphics[width=\linewidth]{stream}
% \caption{Legend (350 words max). Example legend text.}
% \label{fig:stream}
% \end{figure}

% \begin{table}[ht]
% \centering
% \begin{tabular}{|l|l|l|}
% \hline
% Condition & n & p \\
% \hline
% A & 5 & 0.1 \\
% \hline
% B & 10 & 0.01 \\
% \hline
% \end{tabular}
% \caption{\label{tab:example}Legend (350 words max). Example legend text.}
% \end{table}

% Figures and tables can be referenced in LaTeX using the ref command, e.g. Figure \ref{fig:stream} and Table \ref{tab:example}.

\end{document}